%% file: main_arXiv.tex
\documentclass{article}

\usepackage{natbib}

\usepackage{url}            
\usepackage{booktabs}       
\usepackage{amsfonts}       

\usepackage{float}
\usepackage{amsmath}
\usepackage{amssymb}
\usepackage{amsthm}

\usepackage{times}
\usepackage{graphicx} 
\usepackage{subfigure} 
\usepackage{epstopdf}

\usepackage{algorithm}

\usepackage{bbm}
\usepackage[utf8]{inputenc} 
\usepackage[T1]{fontenc}    

\usepackage{nicefrac}       
\usepackage{microtype}      

\usepackage{geometry}
\usepackage{xcolor}
\usepackage{enumitem}
\def\A{\mathcal{A}}
\def\F{\mathcal{F}}
\def\R{\mathbb{R}}
\def\EE{\mathbb{E}}

\def\W{\mathcal{W}}
\def\H{\mathcal{H}}
\def\U{\mathcal{U}}
\def\V{\mathcal{V}}

\def\w{\mathbf{w}}

\def\x{\mathbf{x}}

\def\y{\mathbf{y}}

\def\e{\mathbf{e}}
\def\u{\mathbf{u}}

\usepackage{algorithmic}

\newtheorem{theorem}{Theorem}

\newtheorem{lemma}{Lemma}
\newtheorem{definition}{Definition}

\usepackage{booktabs}

\allowdisplaybreaks

\usepackage{authblk}
\usepackage{hyperref}       

\title{Nonconvex One-bit Single-label Multi-label Learning}

\author[1]{Shuang Qiu\thanks{qiush@umich.edu}}
\author[2]{Tingjin Luo\thanks{tingjinl@umich.edu}}
\author[1,2]{Jieping Ye\thanks{jpye@umich.edu}}
\author[2]{Ming Lin\thanks{linmin@umich.edu}}
\affil[1]{Computer Science and Engineering\\
	University of Michigan, Ann Arbor, MI 48109}
\affil[2]{Department of Computational Medicine and Bioinformatic\\
  University of Michigan, Ann Arbor, MI 48109}

\date{\today}

\begin{document}

\maketitle

\input{lyx_abstract}
\input{lyx_introduction}
\input{lyx_problem}

\input{lyx_algorithm}

\input{lyx_theoretical}

\input{lyx_experiments}
\input{lyx_conclusion}

\small
\bibliographystyle{plainnat}
\bibliography{refs}
\onecolumn
\appendix
\input{lyx_appendix}

\end{document}

%% file: lyx_abstract.tex
\begin{abstract}
We study an extreme scenario in multi-label learning where each training instance is endowed with a single one-bit label out of multiple labels. We formulate this problem as a non-trivial special case of one-bit rank-one matrix sensing and develop an efficient non-convex algorithm based on alternating power iteration. The proposed algorithm is able to recover the underlying low-rank matrix model with linear convergence. For a rank-$k$ model with $d_1$ features and $d_2$ classes, the proposed algorithm achieves $O(\epsilon)$ recovery error after retrieving $O(k^{1.5}d_1 d_2/\epsilon)$ one-bit labels within $O(kd)$ memory. Our bound is nearly optimal in the order of $O(1/\epsilon)$. This significantly improves the state-of-the-art sampling complexity of one-bit multi-label learning. We perform experiments to verify our theory and evaluate the performance of the proposed algorithm.
\end{abstract} 

%% file: lyx_introduction.tex
\section{Introduction}

An important topic in the multi-label learning research is how to exploit the relationship between different classes of labels in order to improve the learning accuracy or reduce the number of required labels. When labels are partially observed, the low-rank matrix model is one of the most popular models to deal with missing labels. As human-labeling is usually expensive and time-consuming, it is critical to design a robust algorithm which is able to learn the underlying low-rank matrix model on datasets with noisy heavily missing labels. In this work, we consider an extreme scenario where each training instance only has one single label being annotated in binary set $\pm 1$ out of multiple classes of labels. This scenario is often encountered in real-world systems but less discussed in literatures. For example, it is rare for a user to annotate a news article or a piece of music with many tags, especially when the user is not paid for his annotation. The problem becomes challenging when we have a large number of features and classes. 

Over the past decades, a number of multi-label learning approaches have been proposed under different settings. Extreme multi-label learning copes with the problem of learning a multi-label classifier from an extremely large scale label set via various machine learning techniques~\citep{bhatia2015sparse, jain2016extreme,xu2016robust,bi2013efficient}. The low rank constraint and the relevant shared structures of the weight matrix are embedded in the loss function in several  previous works~\citep{ji2008extracting,yu2014large,jain2013provable,amit2007uncovering,xu2016robust}. Besides, there are studies trying to tackle the problem of missing labels~\citep{bi2014multilabel, yu2014large} , to provide theoretical analysis on multi-label problem~\citep{jain2013provable} and to deal with multi-label problems via matrix completion~\citep{xu2013speedup, cabral2011matrix}. However, by the time of writing this paper, we are unaware of any work that can cope with our single-label multi-label learning problem in large scale high dimension datasets with provable guarantees.

The contribution of this work is mostly theoretical, although a high performance algorithm is provided as a by-product. We formulate the single-label multi-label learning problem as an one-bit rank-one matrix sensing problem. In our model, the observed label is generated by an rank-one asymmetric matrix sensing operator where the left sensing vector is random Gaussian and the right sensing vector is one-hot random sampling. The parameter matrix is then decomposed as a product of two low-rank unknown matrices, leading to a non-convex learning problem. 

There are several fundamental challenges in our theoretical analysis. The first challenge is the non-convexity of our model. It is hard to argue the global convergence rate from the conventional convex gradient descent framework. Instead, our convergence analysis is built on estimation sequence constructed by the noisy power iteration  \citep{hardt2014noisy}. Our model is a special case of rank-one matrix sensing but with novel structural assumptions. When both left and right sensing vectors have well-bounded sub-gaussian norms or are both one-hot random sampling vectors,  there are several non-convex alternating algorithms with provable guarantees \citep{ zhong2015efficient,hardt2014understanding}. However, in our problem, both assumptions fail to hold true. Our left/right sensing vectors are heterogeneous and our link-function is the signum function which is non-smooth and non-convex. \citet{jain2013provable} once studies a similar sensing operator, but it does not deal with the difficult problem of one-bit signum link-function. These differences make our problem much more challenging than previous rank-one matrix sensing problems.

On the other hand, the one-bit sensing problem has attracted much attention in recent years. There are many emerging insightful works that try to recover a sparse vector from one-bit measurements under Gaussian \citep{plan2013robust,ai2014one,jacques2013robust} or sub-gaussian sensing operators \citep{ai2014one}. \citet{plan2013robust} propose a possible extension of one-bit compressed sensing to matrix sensing where the sensing operator is a standard Gaussian matrix. However, these existing results cannot be directly applied to our problem since our left/right sensing vectors are heterogeneous. The solvers developed in \citep{plan2013robust,ai2014one} are based on convex programming which is less efficient than our alternating updating. \citet{jacques2013robust} propose a gradient descent solver but they require a projection step therefore is difficult to be applied in low-rank problems where the projection step is usually carried out via singular value thresholding.  The most closely related works are recent studies by \citet{bhaskar20151,hsieh2015pu,davenport20141}.  They explore the matrix completion problem under the one-bit setting. However our problem is not a simple matrix completion problem due to the heterogeneity of our left/right sensing vectors. Indeed when reformulated as a matrix completion problem, our problem is ill-proposed thus is not learnable at all. In addition the above works only consider the logistic and least square loss function while we directly incorporate with the signum (hamming) loss function . 

In this paper, we propose a novel non-convex framework to cope with the above challenges with strong theoretical guarantees. We first propose an RIP-type condition between two fixed low-rank matrices under the one-bit setting. Based on the proposed RIP-type condition, we are able to construct an estimation sequence via noisy power iteration. Our method is a gradient-free algorithm  which does not follow the gradient descent direction or minimize the empirical loss function. Given a  model parameter matrix $W\in \mathbb{R}^{d_1 \times d_2}$, our algorithm achieves $\epsilon$ recovery error after retrieving $O(\frac{d_1 d_2 k^{1.5}}{\epsilon})$ one-bit labels. If we apply previous one-bit compressed sensing methods to recover the weight matrix  column by column independently, we will need to retrieve at least $O(\frac{d_1d_2^3}{\epsilon^4})$~\cite{plan2013robust,ai2014one} and $O(\frac{d_1d_2^{1.5}}{\epsilon})$~\cite{jacques2013robust} one-bit labels to achieve the same accuracy. Our bound is significantly better if $k \ll d_2$ .

We organize the remainder paper as following. We present the main problem studied in this paper in Section 2, including the key challenges of our problem and its relation with matrix sensing. In Section 3, we propose a novel non-convex learning algorithm for our problem followed by its theoretical guarantees. The theoretical analysis is sketched in Section 4. Section 5 demonstrates the empirical evaluations of the proposed algorithm. Section 6 concludes this paper. 

%% file: lyx_problem.tex
\section{Problem Formulation}\label{problem}

In this section, we first propose our one-bit low-rank multi-label model. We discuss the main challenges of solving our problem with comparison to the conventional rank-one matrix sensing problems.

\subsection{Notation and Background}
We use $m$, $d_1$, $d_2$ to  denote the number of sampled instances, feature dimensions and the number of labels respectively. Let $W^*\in \R^{d_1 \times d_2}$ be the parameter matrix in our multi-label model. Denote $\x_i \in \R^{d_1}$ to be the feature vector of the $i$-th instance. The feature matrix $X=[\x_1, \x_2,...,\x_m]\in \R^{d_1 \times m}$. The label of the $i$-th instance is  $y_i \in \{-1,1\}$. The label vector $\y = [y_1,...,y_m]^\top$. The one-hot random sampling vector is denoted as $\e_i \in  \R^{d_2}$ where only one element of $\e_i$ is 1 and zero elsewhere. The index of the unique non-zero element of $\e_i$  is uniformly distributed in set $\{1,\cdots, d_2 \}$. The normalized version of $\e_i$ is defined by $\bar{\e}_i = \sqrt{d_2} \e_i$.

In the standard linear regression problem, it is assumed that the responses are generated by a linear function, which is
\begin{align*}
Z = X^T W^*
\end{align*}
where $Z \in \R^{m \times d_2}$ is the matrix of responses in the linear regression. Define the signum function $ \mathrm{sign}(\cdot)$  as 
\begin{align*}
\mathrm{sign}(x) =\begin{cases}
1 & \text{ if } x\geq 0 \\ 
-1 & \text{ if } x < 0 
\end{cases}
\end{align*}
In one-bit multi-label learning, we cannot observe $Z$ directly. Only the sign of $Z$ is observed, that is, $Y_{s,t} = \mathrm{sign}(Z_{s,t}), \forall s \in [m], t \in [d_2]$ where the matrix $Y$ is our observation.  Written in matrix form,
\begin{align}
Y =  \mathrm{sign}(X^\top W^*)
\end{align}
where $\mathrm{sign}(\cdot)$ is generalized to an entry-wise function.

If there is no coorelations between different classes of labels, we have to learn $W^*$ column-by-column as in the conventional one-vs-all classification problem. In particular, it is reasonable to believe that there are underlying correlations between different classes labels such that the labels from one class can be used to improve the estimation of another. By sharing the labels across classes, we might reduce the labeling requirement and improve the quality of our learned model.A popular assumption to capture the correlationship between classes is the low-rank assumption. That is,  we assume $W^*$ to be a low-rank matrix with $\mathrm{rank}(W^*) \leq k$. 

The conventional multi-label learning assumes that all class labels are fully observed. In practice it is usuall difficult to get the full label for all classes. The previous researches model the missing label under matrix completion setting where elements in each row of the label matrix $Y$ are randomly observed. In this paper we consider an even more extreme scenario where only a single class label is retrieved for each training instance. This imposes several novel challenges of recovering $W^*$.  Suppose we retrieve an instance $\x_i$ each time independently where $\x_i$ is a standard random Gaussian vector.  
The observed label $y_i$ is generated via sampling from all the labels of the $i$-th instance by $\bar{\e}_i$, that is
\begin{align} \label{primary}
y_{i} = \mathrm{sign}((\x_i^\top W^*) \bar{\e}_i)  = \mathrm{sign}(\langle \x_i \bar{\e}_i^\top, W^* \rangle)
\end{align}
where $\langle \cdot, \cdot  \rangle$ denotes the matrix inner product. In model Eq. (\ref{primary}), tt is necessary to assume that $W^*$ is a column normalized matrix where each column $W^*_{\cdot, j}$ has a unit norm $||W^*_{\cdot, j}|| = 1$. This is because scaling the input of signum function will not change the value of $y_{i}$. 

We define the set for the matrices of $rank \leq k$ with normalized column as $\F_k$: 
\small
\begin{align*}
\F_k = \{ W \in \R^{d_1 \times d_2} | \mathrm{rank}(W) \leq k, ||W_{\cdot, j}||_2 = 1, \forall j \in [d_2] \}
\end{align*}
Clearly $W^* \in \F_k$. Define the linear operator $\A: \R^{d_1 \times d_2} \mapsto \R^m $ as $\A(W) = [\langle \x_1 \bar{\e}^\top_1, W \rangle,...,\langle \x_m \bar{\e}^\top_m, W \rangle]^\top  \in \R^m$ and $\y = [y_1,y_2,...,y_m]^\top \in \R^m$. Then Eq. (\ref{primary}) can be  equivalently written as 
\begin{align} \label{sensing}
\y =  \mathrm{sign}(\A(W^*))
\end{align}

\paragraph{Remark 1} Note that we do not need to derive any specific optimization objective to solve $W^*$. To estimate $W^*$, we explore an RIP-type condition in Section~\ref{theoretical} and constructed an estimation sequence that converges to $W^*$. The reason for using a scaled sampling vector $\bar{\e}_i$ is to balance the value of $\EE{||\x_i||^2_2}$ and $\EE{||\bar{\e}_i||^2_2}$ such that they are comparable.

\paragraph{Challenges} Our formulation of Eq. (\ref{sensing}) can lead to several key challenges. One of the main challenges is the non-convexity of our problem, which is introduced by  the signum function and the low-rank constraint on the matrix $W^*$. Convex relaxation was a popular choice to cope with non-convex models. Simply dropping the $ \mathrm{sign}(\cdot)$ function and penalize the rank of $W^*$ by its nuclear norm, we can relax our problem as a convex programming problem
$$\hat{W} = \min_W ||\y - \A(W)||_F^2+ \|W\|_*$$
There are at least two drawbacks of the above convex relaxation. First the computation cost of optimizing the nuclear norm regularizer is much more expensive than our non-convex approach based on alternating iteration. Secondly and most importantly, the square loss will introdce the so-called convex bias in learning. It is easy to check that even at $W=W^*$ the above loss function has non-zero gradient. In language of sampling complexity, the square loss will result in $O(1/\epsilon^2)$ sampling complexity which is worse than our nearly optimal $O(1/\epsilon)$ bound.
Another key challenge is the non-smoothness of our problem due to the signum function in Eq. (\ref{sensing}). The theoretically optimal convergence rate for general non-smooth convex optimization is $O(1/\sqrt{t})$ after $t$ iterations. We will show that our alternating iteration will converge linearly, that is exponentially faster than the standard black-box non-smooth convex optimization algorithm. Finally, the non-linearity of $ \mathrm{sign}(\cdot)$ itself can result in a new dimension of challenge in the theoretical analysis. We will encounter this challenge in Section~\ref{theoretical} soon. 

\subsection{Relation with Matrix Sensing}

In matrix sensing, we aim to recover a low rank matrix $W^* (\mathrm{rank}(W^*)\leq k)$ with measurements $\y \in \R^{m}$ generated by some sensing operator $\A : \R^{d_1 \times d_2} \mapsto \R^{m}$. Perhaps the most popular sensing operator is the random Gaussian sensing $\A(W) = [\langle G_1, W \rangle,..,\langle G_i, W \rangle,.., \langle G_m, W \rangle]^\top$, where all the $G_i$'s are the standard Gaussian random matrix. Defining $\y = \mathrm{sign}([\langle G_1, W \rangle,..,\langle G_i, W \rangle,.., \langle G_m, W \rangle]^\top)$, \citet{plan2013robust} generalize the one-bit compressed sensing algorithm in vector space to the  one-bit Gaussian matrix sensing. The method employs the convex relaxation to the low rank constraint using the nuclear norm $\|W\|_*$. The setting is fundamentally different from our setting in this paper. The rank-one matrix sensing problem has been discussed extensively in \citep{zhong2015efficient,hardt2014understanding, jain2013provable}.  In the works \citep{zhong2015efficient,hardt2014understanding}, the sensing operator is $\A(W) = [\langle \u_1 \mathbf{v}_1^\top, W \rangle,...,\langle \u_i \mathbf{v}_i^\top, W \rangle,..., \langle \u_m \mathbf{v}_m^\top, W \rangle]^\top$. Specifically, in their model $\u_i$ and $\mathbf{v}^\top_i$ must be two random vectors sampled from the Gaussian distribution or one-hot random sampling vectors thus is different from our setting.Their $\y$ is not binarized while in our problem, $\y$ is binarized by  $\y =  \mathrm{sign}(\A(W))$. \citet{jain2013provable} studies similar sensing operator as ours, but it does not cope with the challenging problem of binarized $\y$. Therefore we categorize our problem as a new class of  one-bit rank-one matrix sensing. Particularly, the signum function brings the challenge of non-smoothness to our problem that is not studied before in our setting. Moreover, we should note that the alternating minimization algorithm in \citep{zhong2015efficient, jain2013provable} cannot be applied in our one-bit matrix sensing problem since it is difficult to solve the minimization subproblem with $ \mathrm{sign}(\cdot)$ function. Indeed our problem is a non-trivial special case of the one-bit rank-one matrix sensing.

%% file: lyx_algorithm.tex
\section{Algorithm} \label{algorithm} 

\begin{algorithm}[t]
\begin{algorithmic}[1]

\vspace{0.2cm}

\REQUIRE The batch size $m$, batch update iterations
$T$, Rank $k\geq1$. Training instances 
$\y^{(t)}=[y^{(t)}_1,y^{(t)}_2,...,y^{(t)}_m]^T$,
$X^{(t)}=[\x^{(t)}_1,\x^{(t)}_2,...,\x^{(t)}_m]$,
$E^{(t)}=[\bar{e}^{(t)}_1,\bar{e}^{(t)}_2,...,\bar{e}^{(t)}_m]$,$t \in [T]$. And $\lambda =\sqrt{\frac{2}{\pi}}$.

\vspace{0.2cm}

\ENSURE $\U^{(T)},\V^{(T)}$.

\vspace{0.2cm}

\STATE Initialize: $\W^{(0)}=0$, $\V^{(0)}=0$.
$\U^{(0)}=\mathrm{SVD}(\H^{(0)},2k)$, that
is, the top-$k$ left singular vectors.

\vspace{0.2cm}

\FOR{$t=1,2,\cdots,T$}

\vspace{0.2cm}

\STATE Compute $\widetilde{\W}^{(t-1)}= \U^{(t-1)}(\V^{(t-1)})^\top $. 

\vspace{0.2cm}

\STATE  Obtain $\widetilde{W}^{(t-1)}=\widetilde{\W}^{(t-1)}_{[1:d_2, d_2+1:d_2+d_1]}$. 

\vspace{0.2cm}

\STATE Normalize each column $W^{(t-1)}_{\cdot j}= \frac{\widetilde{W}^{(t-1)}_{\cdot j}}{||\widetilde{W}^{(t-1)}_{\cdot j}||}$

\vspace{0.2cm}

\STATE Retrieve $m$ training instances: $\y^{(t-1)}$, $X^{(t-1)}$, and $E^{(t-1)}$.

\vspace{0.2cm}

\STATE $\A(W^{(t-1)})= [...,{\x^{(t-1)}_i}^\top W^{(t-1)} \bar{e}^{(t-1)}_i,...]_{i=1}^m{}^\top$
, and $H^{(t-1)}=\frac{\sqrt{d_2}}{m\lambda}\A'(\y^{(t-1)}-\mathrm{sign}(\A(W^{(t-1)})))$.

\vspace{0.2cm}

\STATE Construct $\H^{(t-1)} =
\begin{bmatrix}
O_{d_2 \times d_1} & H^{(t-1)}\\ 
{H^{(t-1)}}^\top & O_{d_1 \times d_2}
\end{bmatrix}$, and
$\W^{(t-1)} =
\begin{bmatrix}
O_{d_2 \times d_1} & W^{(t-1)}\\ 
{W^{(t-1)}}^\top & O_{d_1 \times d_2}
\end{bmatrix}$.

\vspace{0.2cm}

\STATE $\widetilde{\U}^{(t)}=(\H^{(t-1)}+\W^{(t-1)}) \U^{(t-1)}$.

\vspace{0.2cm}

\STATE Orthogonalize $\widetilde{\U}^{(t)}$ via QR decomposition: $\U^{(t)}=\mathrm{QR}\big(\widetilde{\U}^{(t)}\big)$ .

\vspace{0.2cm}

\STATE $\V^{(t)}=(\H^{(t-1)} + \W^{(t-1)}) \U^{(t)}$

\vspace{0.2cm}

\ENDFOR

\STATE \textbf{Output:} $\U^{(T)},\V^{(T)}$.
\end{algorithmic}

\protect\caption{One-Bit Single-label Multi-label Learning}
\end{algorithm}

In this section, we propose a novel non-convex learning algorithm (Algorithm 1) to recover the $W^*$. We only provide the high level intuition of our algorithm in this section. The rigorous theoretical analysis is postponed to the Section~\ref{theoretical}. 

We first introduce several notations necessary for our analysis. The SVD decomposition of $W^*$ is $W^* = U^* \Sigma^*{V^*}^\top$, where $U^* \in \R^{d_1\times k}$ and $V^* \in \R^{d_2\times k}$. $\Sigma^* = \mathrm{diag}(\sigma_1, ...,\sigma_k)$ is diagonal matrix where $\sigma_1\geq \sigma_2 \geq ... \geq\sigma_k$ are top-$k$ singular values. The adjoint operator of a linear operator $\A(\cdot)$ is $\A'$. We denote $W + O(\epsilon)$ as a matrix $W$ plus a perturbation matrix whose spectral norm is bounded by $\epsilon$.

Our learning problem is non-convex, non-smooth and non-linear. In order to address the three challenges simultaneously, we develop a non-convex learning algorithm based on alternating iteration. Our key idea is to construct an estimation sequence $\{W^{(t)}\}$ to approximate $W^*$. The proposed algorithm is a mini-batch method. In each mini-bathc, it takes $m$ training labels to update $W^{(t)}$. In order to obtain an estimate sequence with reduce variance,  we prove in Theorem~\ref{theorem_diffrip} that 
\begin{align*}\small 
\begin{split}
    W^* = \frac{1}{m} \frac{\sqrt{d_2}}{\lambda} [\A'(\y) - \A'(sign(\A(W^{(t)})))] +  W^{(t)} + O( \delta\tau)
\end{split}
\end{align*}
 provided $W^{(t)} \in \F_k$. The $\tau  = \max  \{ ||W-W'||_2, ||W-W' ||_2^{1/2}, \delta' \}$. When both $||W-W'||_2, ||W-W' ||_2^{1/2}\geq \delta'$ , the perturbation term is $O( \delta\{||W^*-W^{(t)}||_2, ||W-W' ||_2^{1/2}\})$. Intuitively speaking, we construct $W^{(t)}$ to approximate $W^*$ such that the gap $||W^* - W^{(t)}||_2$ shrinks to a small error after sufficient number of iterations. 
 The perturbation term will then decay as $||W^* - W^{(t)}||_2$ gets smaller. 

To simplify our theoretical analysis, inspired by \cite{hardt2014understanding}, we convert the asymmetric matrix  problem into a symmetric one via  Hermitian Dilation techinque. Namely,
\begin{align}\label{newconstruction}
\begin{split}
\W^* = \H^{(t)} + \W^{(t)} + O( \delta||W^*-W^{(t)}||_2)
\end{split}
\end{align}
where we define 
\begin{align*}
&\W^* = \begin{bmatrix}
O_{d_2 \times d_1} & W^*\\ 
{W^*}^\top & O_{d_1 \times d_2}
\end{bmatrix},
\W^{(t)} =\begin{bmatrix}
O_{d_2 \times d_1} & W^{(t)} \\ 
W^{(t)}{}^\top & O_{d_1 \times d_2} \\
\end{bmatrix},
\H^{(t)} = \begin{bmatrix}
O_{d_2 \times d_1} & H^{(t)} \\ 
H^{(t)}{}^\top & O_{d_1 \times d_2} ~.
\end{bmatrix}
\end{align*}
In the above $H^{(t)} = \frac{1}{m} \frac{\sqrt{d_2}}{\lambda} [\A'(\y) - \A'(sign(\A(W^{(t)})))]$ and $O_{d_1 \times d_2} \in \R^{d_1 \times d_2}, O_{d_2 \times d_1} \in \R^{d_2 \times d_1}$ are two zero matrices. Specifically, $\W^*$ has a rank of $2k$ and singular values $\sigma_1,...,\sigma_k$ each occuring with multiplicity two. More detailed properties of this symmetric construction is presented in the next section. By construction, the  estimation sequence in Eq. (\ref{newconstruction}) will converge to $\W^*$ if and only if $W^{(t)}$ converge to  $W^*$. More precisely, $\W^{(t)}$ is obtained by updating two parameter matrix $\U^{(t)}, \V^{(t)} \in \R^{(d_1+d_2) \times 2k}$ alternatively. In line 3-5 of Algorithm 1, denoting $\widetilde{\W}^{(t)} = \U^{(t)} \V^{(t)}{}^\top$, we can extract the block matrix of row $1$ to $d_2$, column $d_2+1$ to $d_2 + d_1$ from $\widetilde{\W}^{(t)}$, perform column normalization to get $W^{(t)}$, and then build the symmetric matrix $\W^{(t)}$ as in Eq. (\ref{newconstruction}).  Finally, we export $W^{(T)}$ by the learned $\U^{(T)}$, $\V^{(T)}$.

The two parameter matrix $\U^{(t)}$, $\V^{(t)}$ require space complexity of $O(k(d_1+d_2))$. Other related variables $W^{(t)}$, $H^{(t)}$, $\W^{(t)}$, $\H^{(t)}$ can be computed from $\U^{(t)}$,$\V^{(t)}$ oon-the-fly. During one mini-batch updating, only inner product operations are required, which can be efficiently implemented on many computation architectures. The algorithm is initialized via truncated SVD which can be done via power iteration. The QR step on requires $O(k^2 (d_1+d_2))$ computing complexity, which is more efficient than SVD when $k \ll d_1+d_2$. Algorithm 1 retrieves instances in stream, a favorable behavior on systems with high speed cache.

The main theoretical result is presented in the following theorem, which gives the convergence rate of recovery and sampling complexity for our problem. The proof of this theorem is postponed to the end of the next section.

\begin{theorem}\label{theorem_main}
Suppose $\x_i$ and $\bar{\e}_i$ are i.i.d. sampled. $W^*$ is a rank-$k$ matrix. Then with probability at least $1-\eta$, there exists a constant $C$, a constant $\delta_1 \leq 1$ and $t_0 > 0$ such that
\begin{align}
||W^* - W^{(t+1)}||_2 \leq \delta_1^t ||W^*||_2
\end{align}
when $t < t_0$, and
\begin{align}
||W^* - W^{(t+1)}||_2 \leq \delta_1^{2-2^{t_0-t}} ||W^{(t_0)}-W^*||_2^{2^{t_0-1-t}}
\end{align}
when $t \geq t_0$,
provided 
\[m \geq C (12\sqrt{5} \sigma_1^*/ \sigma_k^* + 4)^2 d_1 d_2 k^{1.5} / \delta_1^2 ~.\] 
The $t_0$ is the smallest integer such that $||W^* - W^{(t0)}||_2 \leq 1$.
\end{theorem}

Theorem ~\ref{theorem_main} demonstrates that ${W^{(t+1)}}$ converges to $W^*$ linearly at the beginning when the gap $||W^{(t+1)}- W^*||_2 > 1$, which is controlled by $\delta_1$. But after $W^*$ and $W^{(t+1)}$ are sufficiently close, $||W^{(t+1)}- W^*||_2 \leq 1$ then the convergence rate becomes $\delta_1^{2-2^{t_0-t}}$. Note that $\delta_1$ is a constant of order $O(\sqrt{d_1 d_2 k^{1.5}/m})$. It is easy to check that if $t \rightarrow \infty$, $||W^* - W^{(t+1)}||_2 \leq \delta_1^2 = O(d_1d_2k^{1.5}/m)$, which indicates $W^{(t+1)}$ can reach a point $O(1/m)$ far from $W^*$ as $m$ increases. We further have 
\begin{align*}
&\delta_1^{2-2^{t_0-t}} ||W^{(t_0)}-W^*||_2^{2^{t_0-1-t}} \\
= &\delta_1^2 (||W^{(t_0)}-W^*||_2/\delta_1^2)^{2^{-t+t_0-2}}\\
= &\delta_1^2 \omega_{t_0}^{2^{-t+t_0-2}} ~.
\end{align*}
Note that $\omega_{t_0} \geq 1$ when $t > t_0$. Therefore within a limit number of $T$ iterations,  when $T > t_0$, $\omega_{t_0}^{2^{-T+t_0-2}}$ will decay so fast that $\delta_1^2 = O(d_1d_2k^{1.5}/m)$ dominates the recovery error. This indicates Algorithm 1 achieves $O(\epsilon)$ recovery error after retrieving $mT = O(d_1d_2k^{1.5}/\epsilon)$ instances. Additionally, a small $\delta_1$ will result in a fast convergence rate as well as a small recovery error but a large sampling complexity. The sampling complexity is of the order $O((12\sqrt{5} \sigma_1^*/ \sigma_k^* + 4)^2 d_1 d_2 k^{1.5} / \delta_1^2)$. The sampling complexity is controlled by the condition number $\sigma_1^*/ \sigma_k^*$. The dependence on the condition number can be removed via the soft-deflation trick when the singular values decrease fast enough.

\noindent\textbf{Comparison.} Previous one-bit sensing researhes mainly focused on the one-bit compressed sensing in sparse vector space. There is rare work to study the 1-bit problem in low rank matrix space. For the comparisons, we think of applying one-bit compressed sensing methods to recover the weight matrix $W^{d_1 \times d_2}$ column by column, to analyze their sampling complexity. And we assume all the columns of $W$ is not sparse. By the papers~\cite{plan2013robust,ai2014one}, it requires $\frac{d}{\epsilon^4}$ to recover a non-sparse $d$-dimensional vector to error $O(\epsilon)$. And By the paper~\cite{jacques2013robust}, it requires $\frac{d}{\epsilon}$ for $\epsilon$ recovery, which is much better. However, in our problem, to guarantee recover the matrix $W$ with error $O(\epsilon)$, we need those algorithms to recover each column with error $\epsilon/\sqrt{d_2}$. And the overall sampling complexity is $d_2$ times of samples for each single column. It is not difficult to show that it requires $O(\frac{d_1d_2^3}{\epsilon^4})$~\cite{plan2013robust,ai2014one} and $O(\frac{d_1d_2^{1.5}}{\epsilon})$ training instances by ~\cite{jacques2013robust} to achieve overall $O(\epsilon)$ recovery error and $O(\frac{d_1d_2^{1.5}}{\epsilon})$ training instances by ~\cite{jacques2013robust}. However, our method can reduce it to only $O(\frac{d_1 d_2 k^{1.5}}{\epsilon})$, which achieves an improvement if $k \ll d_2$ in a real scenario. Note that we omit the $\log$ terms in our analysis.

%% file: lyx_theoretical.tex
\section{Theoretical Analysis}\label{theoretical}

In this section, we present the necessary lemmas and theorems to build a proof structure for the convergence of Algorithm 1. And at the end of this section, we demonstrate the proof of the Theorem~\ref{theorem_main}.

As introduced in Section~\ref{problem} and Section~\ref{algorithm}, the main idea of our proposed algorithm is to construct estimation sequence $W^{(t)}$ such that this sequence can eventually approximate $W^*$ with a tiny approximation error. 

Before presenting our theoretical analysis, we introduce an definition known as the restricted isometry property (RIP)~\cite{candes2012exact}.
\begin{definition}[Restricted Isometry Property]
A linear sensing operator $\A$ satisfies $\delta_k$-RIP if for any rank $k$ matrix $W$, 
$$(1-\delta_k) ||W||_F^2 \leq \frac{1}{m} \leq (1+\delta_k) ||W||_F^2$$
where $\delta_k \in (0,1)$.
\end{definition}
This conventional RIP condition cannot be applied to our analysis here since this RIP condition is not able to take the one-bit binarization into consideration. Therefore, this provides little help to construct an estimation sequences as we expect. \citet{lin2016non} resorts to proposing an RIP-type condition with $||\frac{1}{m} \A'\A(W) - \EE \{ \A'\A(W)\}||_2 \leq \delta ||W||_2$ such that an estimation sequence is built by replacing $W = W^* - W^{(t)}$ and then $\A'\A(W)$ becomes $\A'\A(W^*) - \A'\A(W^{(t)})$ due to the linearity of the operator $\A$ and the associated adjoint operator $\A'(\cdot)$. However, this idea is also not completely fit for our problem because we are incapable of obtaining any real-value magnitude of $\A(W^*)$ due to signum function. Instead, we consider to explore the possibility of designing RIP-type condition using $\A' \mathrm{sign}(\A(W))$. Unfortunately, the signum function brings a great difficulty to replace $W$ by $W^* - W^{(t)}$ due to its non-linearity and $\A' \mathrm{sign}(\A(W^* - W^{(t)})) = \A' \mathrm{sign}(\A(W^*)) - \A' \mathrm{sign}(W^{(t)})$ does not hold mathematically. However, we successfully tackle this problem by finding a RIP-type condition in another form. We discover that we can directly analyze $\A' \mathrm{sign}(\A(W^*)) - \A' \mathrm{sign}(\A(W^{(t)}))$ and its associated expectation instead of using the previous formulations. 

We can derive our bound based on matrix Bernstein's inequaltiy. Thus, the expectation of the term $\A' \mathrm{sign}(\A(W^*)) - \A' \mathrm{sign}(\A(W^{(t)}))$ is required to compute. The signum function in this term can bring a lot of difficulties to compute the associated expectation. In Lemma~\ref{lemma_expectation}, we show an approach to compute the expectation of $\A' \mathrm{sign}(\A(W))$ with a mathematical proof. Although there exists a signum function in the term below, we still succeed to obtain the expectation value.
\begin{lemma}\label{lemma_expectation}
Let $X = [\x_1,...,\x_m] \in \R^{d_1 \times m}$ be a matrix with i.i.d. standard Gaussian entries. And let $\x_i \bar{\e}_i^\top$ be the measurement operator defined by Eq.~\ref{primary}. For a matrix $W \in \R^{d_1 \times d_2}$ with $||W_{\cdot j}||_2 = 1, \forall j \in [d_2]$, we have
\begin{align*}
\EE\{ \mathrm{sign}(\langle  \x_i \bar{\e}_i^\top, W \rangle)  \x_i \bar{\e}_i^\top\} = &\frac{\lambda}{\sqrt{d_2}} W
\end{align*}
where $\lambda$ is a constant $\lambda = \sqrt{\frac{2}{\pi}}$. Furthermore, letting $\A : \R^{d_1\times d_2} \mapsto \R^m$ be the sensing operator as in Eq.~\ref{sensing} and $\A': \R^m \mapsto \R^{d_1\times d_2}$ be the associated adjoint operator, we have
\begin{align*}
\frac{1}{m}\EE\{\A'( \mathrm{sign}(\A(W)\} = \frac{\lambda}{\sqrt{d_2}} W
\end{align*}
\end{lemma}
Please refer to Appendix for detailed proof. Under the assumption of standard Gaussian distributions of $\x_i$, the Lemma~\ref{expectation} shows that we obtain a linear relation between $\EE \A' \mathrm{sign}(\A(W))$ and $W$. 

Furthermore, we prove one of the most important lemmas in our framework as below. This lemma is quite critical for us to propose our RIP-type condition in in Theorem~\ref{theorem_diffrip}, since it manages to extract the relation of $||\w-\w'||_2$ from the inside of two signum functions in the formulation below. And the term $||\w-\w'||_2$ will finally lead to the term  $||W^*-W^{(t)}||_2$ in Theorem~\ref{theorem_diffrip}. This plays an important role in constructing a global convergent estimation sequence.

\begin{lemma}\label{lemma_distance}
Let $ \boldsymbol g \in \R^{d_1}$ be a vector with i.i.d. standard Gaussian entries. For two different vectors $\w, \w' \in \R^{d_1}$, if $\arccos(\langle \w,\w' \rangle) \leq \frac{\pi}{2}$, then we have
\begin{align*} 
    &||\EE\{\boldsymbol g \boldsymbol g^\top | \mathrm{sign}(\langle \boldsymbol g , \w \rangle) -  \mathrm{sign}(\langle \boldsymbol g , \w' \rangle) |^2 \}||_2 \leq C_1 ||\w - \w'||_2 \\
    &||\EE\{ ||\boldsymbol g||_2^2 | \mathrm{sign}(\langle \boldsymbol g , \w \rangle) -  \mathrm{sign}(\langle \boldsymbol g , \w' \rangle) |^2 \} ||_2 \leq C_2 d_1 ||\w - \w'||_2 
\end{align*}
\end{lemma}
The proof of this lemma is presented in Appendix. The key challenge of proving this lemma is to construct a rotation matrix applied to $\w, \w'$ and $\boldsymbol g$ to transform the our proof from a $d_1$-dimensional space into $2$-dimensional space. We find that the terms $||\EE\{\boldsymbol g \boldsymbol g^\top | \mathrm{sign}(\langle \boldsymbol g , \w \rangle) -  \mathrm{sign}(\langle \boldsymbol g , \w' \rangle) |^2 \}||_2$ and $||\EE\{ ||\boldsymbol g||_2^2 | \mathrm{sign}(\langle \boldsymbol g , \w \rangle) -  \mathrm{sign}(\langle \boldsymbol g , \w' \rangle) |^2 \} ||_2$ are rotation-invariant. And then with integration and some properties of standard Gaussian distribution, we can finally prove this lemma. 

The following theorem present the RIP-type condition for our problem based on the above lemmas. 
\begin{theorem}\label{theorem_diffrip}
Suppose $\A(\cdot) = [\langle \x_1 \bar{\e}_1^\top, \cdot \rangle, ...,\langle \x_m \bar{\e}_m^\top, \cdot \rangle]^T$ are defined as in Eq.\ref{sensing}. $W \in \R^{d_1 \times d_2}$ and $W' \in \R^{d_1 \times d_2}$ are two different column normalized fixed matrices, $d_1 > d_2$, $W \neq W'$. Then with a probability at least $1-\eta$, provided $m \geq C d_1 d_2 k^\frac{1}{2} / \delta^2$ and $\delta' = \delta / (Cd_1^{1/2}k^{1/2})$
\begin{align*}
||\frac{\sqrt{d_2}}{\lambda m} \bigg(\A' ( \mathrm{sign}(\A(W))) - \A'(\mathrm{sign}(A(W')))\bigg) - (W-W')||_2 \leq \delta \max  \{ ||W-W'||_2, ||W-W' ||_2^{1/2}, \delta' \}
\end{align*}
where $\delta$ is of the order $O(\sqrt{d_1 d_2 k^{1/2}/m})$.
\end{theorem}
Please refer to Appendix for the detailed proof. By this theorem, we let $W = W^* \in \F_k$ and $W'=W^{(t)} \in F_{2k}$ and then we can obtain the similar result according to this theorem. Using Bernstein's Inequality, we can also extend this theorem to the case where $W' = 0$ and $W = W^*$, which is right the initialization process of Algorithm 1. 

Symmetric matrix is much easier to analyze than a general matrix. Hermitian Dilation is a quite useful technique, which can construct a symmetric matrix of rank $2k$ using a general rank-$k$ matrix while keeping its original spectral information.  

\begin{lemma}[Hermitian Dilation~\cite{tropp2015introduction,zhang2015singular}]\label{lemma_hermitian} Suppose $W \in \R^{d_1 \times d_2}$ is a matrix of rank $k$ whose SVD is $U\Sigma V^\top$ with $\Sigma = diag(\sigma_1,...,\sigma_k)$. The Hermitian dilation is a map from a general matrix to a Hermitian matrix defined by
\begin{align*}
    \W = \begin{bmatrix}
O_{d_2 \times d_1} & W \\ 
{W}^\top & O_{d_1 \times d_2}
\end{bmatrix}
\end{align*}
And the SVD of $\W$ is 
\begin{align*}
\W = \begin{bmatrix}
U/\sqrt{2} & U/\sqrt{2} \\ 
V/\sqrt{2} & -V/\sqrt{2}
\end{bmatrix} 
\begin{bmatrix}
\Sigma & O_{k \times k} \\ 
O_{k \times k} & -\Sigma
\end{bmatrix}
\begin{bmatrix}
U/\sqrt{2} & U/\sqrt{2} \\ 
V/\sqrt{2} & -V/\sqrt{2}
\end{bmatrix}^\top
\end{align*}
which indicates that $||\W||_2 = ||W||_2$, the eigen values of $\W$ take the values of $\pm \sigma_i, \forall i \in [k]$, and $rank(\W) = 2\cdot rank(W) = 2k$.  
\end{lemma}
Lemma~\ref{lemma_hermitian} shows that the rank of the dilated matrix becomes twice as the original one while the spectral norm remains the same. We use this important property through the below analyses. 

Based on Lemma~\ref{lemma_hermitian}, we can assume $\W^* = \U^* \Lambda^* \U^*{}^\top$ is the eigenvalue decomposition of $\W^*$. Therefore, in Algorithm 1, we resort to construct the estimation sequence $\W^{(t)}$ to estimate $\W^*$. We first use  $\widetilde{\W}^{(t)}= \U^{(t)}\V^{(t)}{}^\top$. And then we extract $\widetilde{W}^{(t)}$ from $\widetilde{\W}^{(t)}$ and perform column normalization to get $W^{(t)}$ which further leads to $\W^{(t)}$. Utilizing the Hermitian Dilation, we can finally show that $||W^* - W^{(t)}||_2$ converges with $||\W^* - \W^{(t)}||_2$ converging.

By the Theorem~\ref{theorem_diffrip}, we can have the following relation 
\begin{lemma} \label{lemma_sequence}
Let $W^{(t)}$, $H^{(t)}$, $\W^{(t)}$, $\H^{(t)}$ be defined as in Algorithm 1. And let $\epsilon^{(t)} = ||W - W^{(t)}||_2$. Then with a probability at least $1-\eta$, provided $m \geq C k^{1/2} d_1 d_2 / \delta^2 $ and $\delta' = c \delta / (d_1^{1/2}k^{1/2})$,
\begin{align*}
W^* + O(\delta \tau) =  H^{(t)} + W^{(t)} 
\end{align*}
And by the Hermitian dilation, we further obtain
\begin{align*}
 \W^* + O(\delta \tau) =  \H^{(t)} + \W^{(t)}     
\end{align*}
where $\tau = \max  \{ ||W-W'||_2, ||W-W' ||_2^{1/2}, \delta_1 \}$.
\end{lemma}
This can be directly shown by Theorem~\ref{theorem_diffrip}. We omit the proof here.

\begin{lemma}\label{lemma_loss}
Suppose $\W^*$ is the Hermitian Dilation of $W^*$, and $\W^{(t)}$ are constructed via $\widetilde{\W}^{(t)}$ as in Algorithm 1. Then we have 
\begin{align*}
    ||\W^* - {\W}^{(t)} ||_2 \leq 4\sqrt{k}||\W^* - \widetilde{\W}^{(t)} ||_2
\end{align*}
\end{lemma}
In each iteration, we need to extract $\widetilde{W}^{(t)}$ from $\widetilde{\W}^{(t)}$ and perform column normalization as well as Hermitian Dilation to obtain $\W^{(t)}$. The error increment caused by this process can be bounded in Lemma~\ref{lemma_loss} as we prove. It is helpful for us to analyze the convergence of our algorithm.

The following lemma shows the canonical angle between column spaces of two different matrices. And in our problem setting, we need to measure the angle between the two subspaces of $\U^*$ and $\widetilde{\U}^{(t)}$ such that our algorithm converges if this angle can converge. 
\begin{lemma}\label{lemma_angle}
Let $\theta(\U^*, \widetilde{\U}^{(t)})$ be the largest canonical angle to measure the distance of two subspaces that are  respectively spanned by $\U^*$ and $\widetilde{\U}^{(t)}$. Supposing $\U^{(t)} R = \widetilde{\U}^{(t)}$ is the QR decomposition step of Algorithm 1, we have
\begin{align*}
\tan \theta(\U^*, \widetilde{\U}^{(t)})  = \tan \theta(\U^*, {\U}^{(t)})  
\end{align*}
where $\U^{(t)} \in \R^{d \times 2k}$ has orthonormal columns and $R \in \R^{2k \times 2k}$ is upper triangular. If $\widetilde{\U}^{(t)}$ is full-rank, so are $\U^{(t)}$ and $R$.
\end{lemma}

We consequently define $\theta_t \triangleq \theta(\U^*, \widetilde{\U}^{(t)}) = \theta(\U^*, {\U}^{(t)})$, $\alpha_t \triangleq \tan \theta_t$, $\epsilon_t \triangleq ||\W^* - \W^{(t)}||_2$.
\begin{lemma}\label{lemma_descent}
Under the same setting of Theorem~\ref{theorem_diffrip}, suppose $\alpha_t \leq 2$ and $\max\{\epsilon_t, \epsilon_t^{1/2}\} \leq \frac{2}{3\sqrt{5}} \sigma_k^* /\delta$ and both $\epsilon_t, \epsilon_t^{1/2} \geq \delta'$, then 
\begin{align*}
    &\alpha_{t+1} \leq 3\sqrt{5} \delta \sigma_{k}^*{}^{-1} \max\{\epsilon_t, \epsilon_t^{1/2}\}\\
    &\epsilon_{t+1} \leq 4\sqrt{k} \alpha_{t+1} \sigma^*_1 + 4\sqrt{k} \delta \max \{\epsilon_{t},\epsilon_{t}^{1/2} \}
\end{align*}
\end{lemma}
We can use Lemma~\ref{lemma_descent} to prove that our algorithm can converge to a very tiny error under the condition in the lemma. The recursion relation in the above formulation can lead to the convergence.

Then we need further discuss the condition to guarantee that the initial value $\alpha_0$ satisfies the assumption of Lemma~\ref{lemma_angle}. Then, we need the following lemma which directly applies Wely’s and Wedin’s theorems ~\cite{stewart1990matrix}. 
 
\begin{lemma}\label{lemma_pert}
Denote $\U \in \R^{d\times 2k}$ and $\U' \in \R^{d\times 2k}$ as the top-$2k$ left singular vectors of $\W$ and $\W' = \W + O(\epsilon)$ respectively. The i-th singluar value of $\W$ is $\sigma_i$. Suppose that $\epsilon \leq \frac{\sigma_{2k} - \sigma_{2k+1}}{4}$. Then the largest canonical angle between the subspaces spanned by $\U$ and $\U'$ is bounded by $\sin \theta(\U, \U')\leq 2\epsilon /(\sigma_{2k} - \sigma_{2k+1})$
\end{lemma}

According to Lemma~\ref{lemma_pert}, when $\delta \sigma_1 = \delta \sigma^*_1\leq \sigma_{2k}/4 = \sigma^*_{k}/4 $, we have $\sin \theta_0 \leq 2 \delta \sigma^*_1 / \sigma^*_k$. Therefore, $\alpha_0 \leq 2$ provided $\delta \leq \sigma^*_k / (4\sigma^*_1)$. And thus we find the condition for $\alpha_0 \leq 2$.

\subsection{Proof of Theorem~\ref{theorem_main}}

Under the setting of Theorem~\ref{theorem_diffrip}, we can use the inequality in it.
\begin{proof}
By lemma~\ref{lemma_descent}, we can have
\begin{align*}
    \epsilon_{t+1} &\leq 4\sqrt{k}(\alpha_{t+1} \sigma^*_1 + \delta \max\{\epsilon_t, \epsilon_t^{1/2}\})\\
    &\leq 4\sqrt{k}(3\sqrt{5}\delta  \sigma^*_1 / \sigma_k^* \max\{\epsilon_t, \epsilon_t^{1/2}\} + \delta \max\{\epsilon_t, \epsilon_t^{1/2}\})\\
    &=\sqrt{k}(12\sqrt{5}  \sigma^*_1 / \sigma_k^*  + 4 )\delta \max\{\epsilon_t, \epsilon_t^{1/2}\}
\end{align*}

Now we can split our proof into two cases where $\epsilon_t > 1$ and $\epsilon_t \leq 1$. Clearly, $\epsilon_t > 1 \Leftrightarrow \epsilon_t > \epsilon_t^{1/2}$ and $\epsilon_t \leq 1 \Leftrightarrow \epsilon_t \leq \epsilon_t^{1/2}$

When $\epsilon_t > 1$, we have
\begin{align*}
\epsilon_{t+1} \leq (12\sqrt{5}  \sigma^*_1 / \sigma_k^*  + 4 )\sqrt{k} \delta \epsilon_t
\end{align*}
which indicates that
\begin{align*}
\epsilon_{t+1} \leq ((12\sqrt{5}  \sigma^*_1 / \sigma_k^*  + 4 )\sqrt{k} \delta)^t \epsilon_0
\end{align*}
Let $\delta_1 = (12\sqrt{5}  \sigma^*_1 / \sigma_k^*  + 4 )\sqrt{k} \delta$. Only $\delta_1 < 1$ can guarantee the convergence, which implies,
\begin{align*}
 \delta < \frac{\sigma_k^*}{(12\sqrt{5}  \sigma^*_1   + 4 \sigma_k^*)\sqrt{k}}
\end{align*}

Suppose after $t_0$ iterations, $\epsilon_t$ decreases to $\epsilon_{t_0} \leq 1$. Let $\delta_1 < 1$. And now we have, for $t \geq t_0$
\begin{align*}
\epsilon_{t+1}\leq \delta_1 \epsilon_t^{1/2}
\end{align*}
which implies that the error $\epsilon_{t+1}$ can be 
\begin{align*}
    \epsilon_{t+1}&\leq \delta_1^{1+\frac{1}{2}+...+\frac{1}{2^{t-t_0}}} \epsilon_{t_0}^{\frac{1}{2^{t+1-t_0}}} \\
    & = \delta_1^{2-\frac{1}{2^{t-t_0}}} \epsilon_{t_0}^{\frac{1}{2^{t+1-t_0}}}
\end{align*}
Note that the smallest possible value of $\epsilon_{t_0}$ is $\delta_1$ which naturally satisfies $ \delta_1 \geq \delta_1^2$ if $\delta_1 < 1$. And thus $\epsilon_{t_0} \geq \delta_1^2$ can always hold, which is able to guarantee $\epsilon_t$ decreasing when $t \geq t_0$, which means $\epsilon_{t+1} \leq \epsilon_t$
Since $\epsilon_{t_0} \leq 1$, then with $t\rightarrow +\infty$, there is
$$
\epsilon_{t+1}\leq  \delta_1^2
$$
which shows our algorithm can globally converge to a small error $\delta_1^2$ in an order of $O(\frac{d_1d_2k^{1.5}}{m})$. And thus $\delta_1^2 > \delta'$. By the property of Hermitian Dilation in Lemma~\ref{lemma_hermitian}, $||W^* - W^{(t+1)} ||_2 = ||\W^* - \W^{(t+1)} ||_2 = \epsilon_{t+1}$. Thus $||W^* - W^{(t+1)} ||_2 \leq \delta_1^2$ as $t \rightarrow \infty$.This completes the proof. 
\end{proof}
From the proof, we obtain that it requires $\delta < \frac{\sigma_k^*}{(12\sqrt{5}  \sigma^*_1   + 4 \sigma_k^*)\sqrt{k}}$ which can guarantee $\alpha_t < 2$ for $\forall t \geq 0$.

%% file: lyx_experiments.tex
\section{Experiments}
In this section, we empirically verify our method on the synthetic datasets for two different situation. In Subsection~\ref{streaming}, we implement our algorithm for the case where each training instance only has single label for training, to verify the validity of our theory that is mainly studies in this paper. In Subsection~\ref{streaming}, we turn to extend our method to the multi-label learning with full observations, which is based on conventional assumption that all the labels for training instance are observed and there is no missing labels. And we demonstrate the way to extend our algorithm in this subsection. Our algorithm is implemented with MATLAB 2016b version. Our computer has 64 GB memory and a 64 bit, 16 core CPU. 

\subsection{Streaming Data With Single Feedback}\label{streaming}
In this subsection, we implement our algorithm for the streaming instances with only single one bit label feedbacks. In our paper, we assume that the instances are generated in a streaming way and there is randomly only one label can be observed. Our implemented algorithm can collect a batch of streaming data each iteration and use them for training only one update of the parameters. And then it collects another batch of training instances. 

In our experiments, we construct the $W^*$ of rank $k$ by two rank $k$ matrix $U \in \R^{d_1 \times k}$ and $V \in \R^{d_2 \times k}$, whose elements are generated by standard Gaussian distribution. And we perform column normalization to $U V^\top$ to obtain $W^*$. In order to train our model, in each iteration, we generate m instances $\x_i$ whose each entry follows i.i,d standard Gaussian $\mathcal{N}(0,1)$, and the single training label for one instance is randomly sampled from the vector $W^*{}^\top \x_i$. And then we test our model based on a batch of new instances and their whole labels. 

To further explore the effect of noisy data to our learning algorithm, we add two kind of noises to the training data. (1) Adversarial Noise. We randomly flip a small percentage $p$ of training labels and use them to train our model.(2) Random noise. We obtain the training labels by adding a small perturbation to the generating process by $y_i =\mathrm{sign}(\langle \x_i \bar{\e}_i^\top, W^* \rangle + \boldsymbol \xi)$ where $\xi$ is a Gaussian random variable with small variance $\mathcal{N}(0,\xi^2)$.

We compare our algorithm with a state-of-art method, LEML\cite{yu2014large}, which is a low rank multi-label method, proposed to solve multi-label problems with missing data in high efficiency and accuracy. It is able to be extended to our extreme case of only single label is observed.  In our experiment, we performed the two algorithm for 10 iterations and compare the methods under the cases of free noise and two different noises respectively. We set $p = 1\%, 2.5\%, 5\%$ and $\xi = 0.1, 0.2, 0.3$. Additionally, we set $d_1 = 500$, $d_2 = 200$, $k = 3$ and $m = 100000$ for training in each iteration. And then we test our model by $10000$ instances with their overall labels. For the evaluation metric, we apply the prediction error defined as the Hamming loss and the average AUC. 

\begin{figure*}[t]
	\centering
	\subfigure[Noise Free]{
		\includegraphics[width = 0.23 \textwidth]{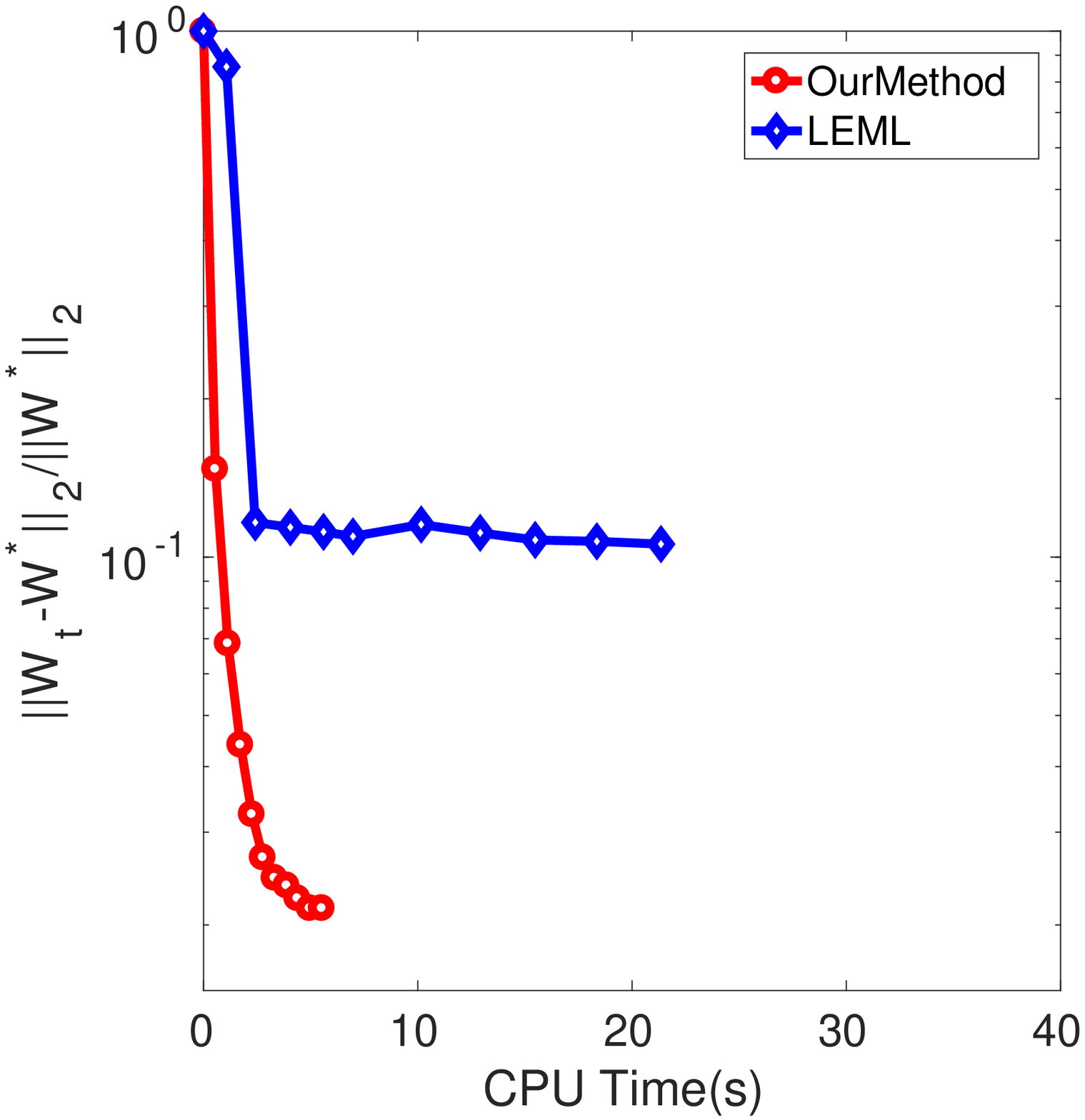}}
	\subfigure[Noise Free]{
		\includegraphics[width = 0.24 \textwidth]{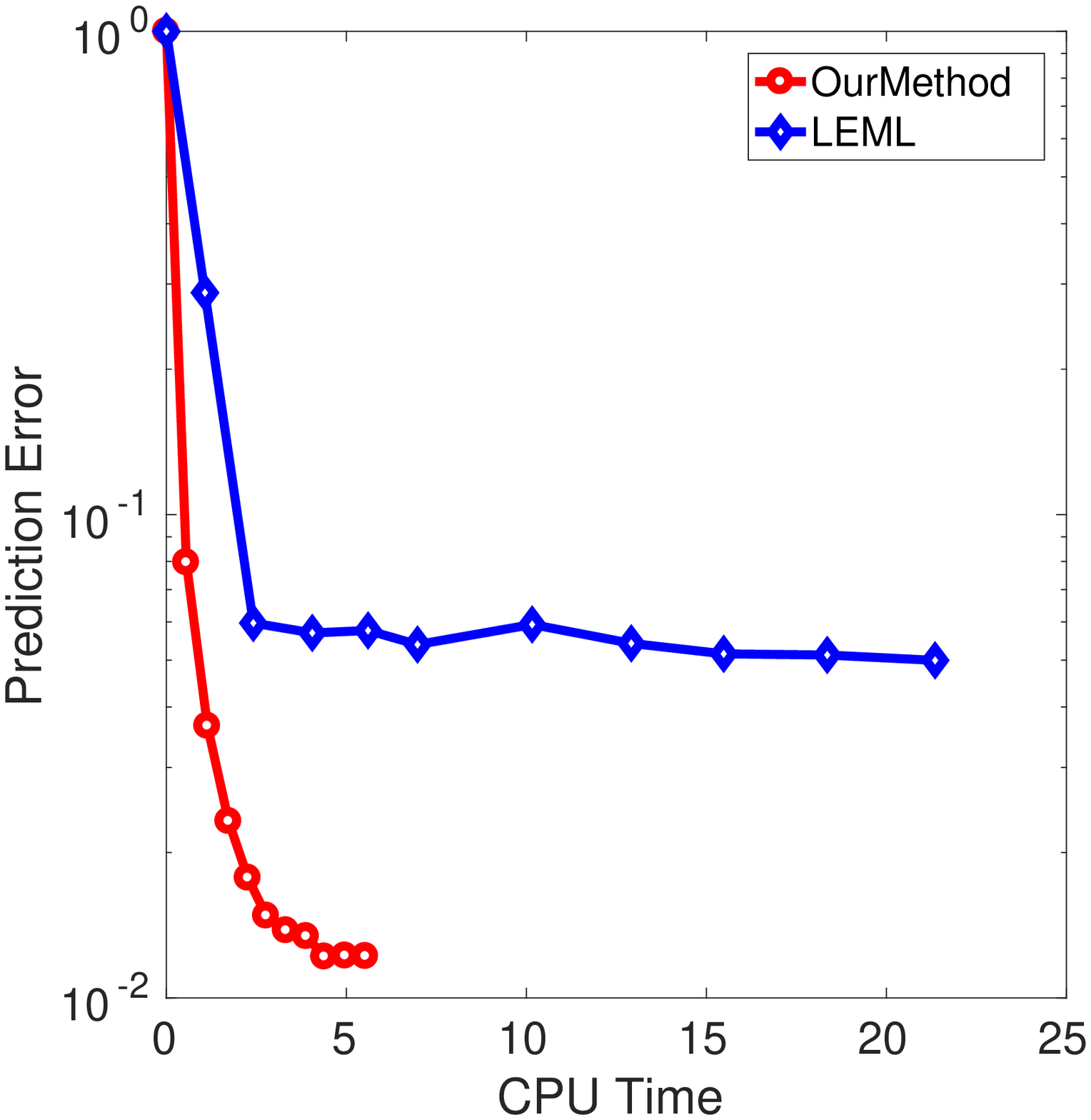}}
	\subfigure[$\xi = 0.1$]{
		\includegraphics[width = 0.24 \textwidth]{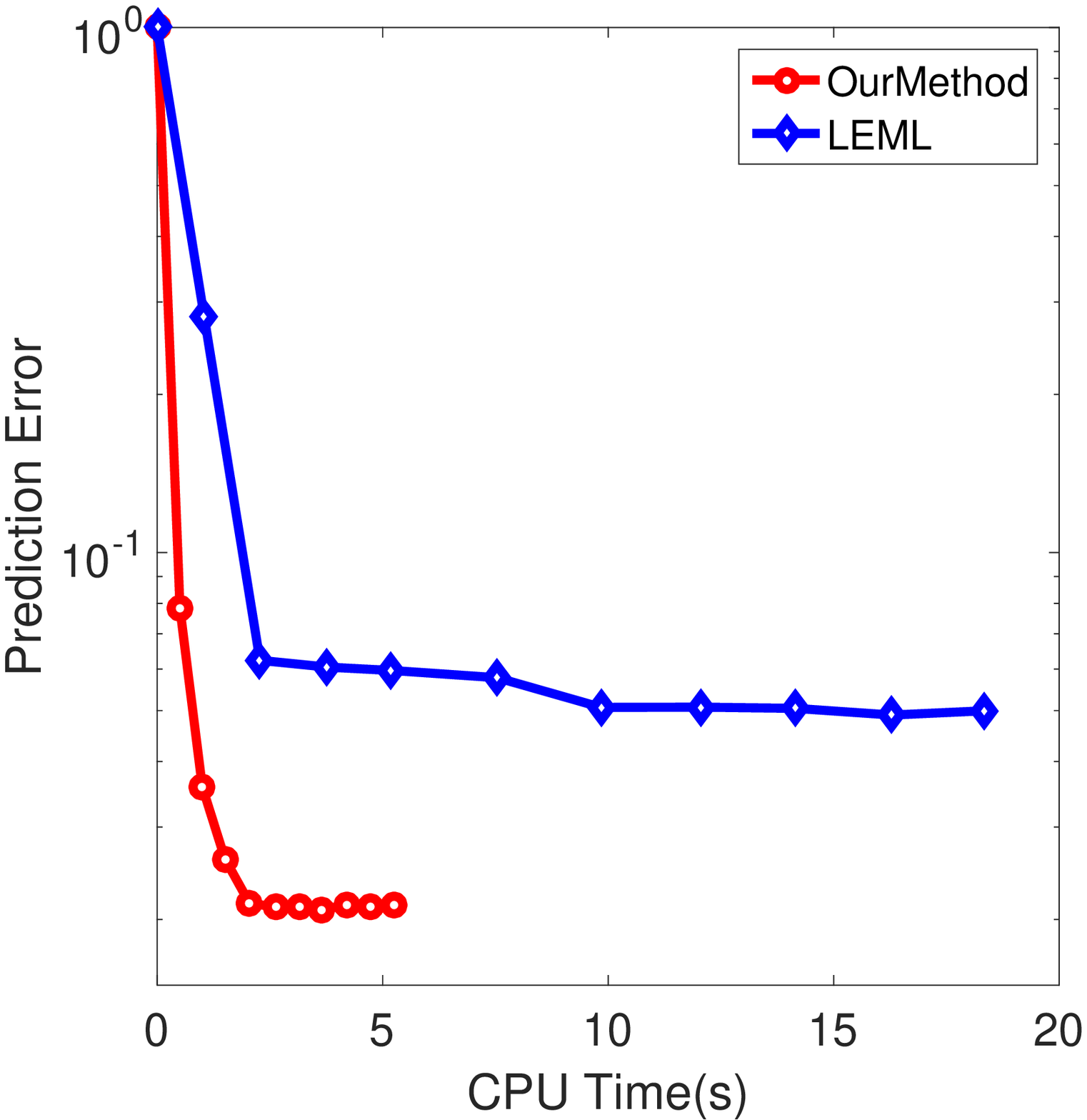}}
	\subfigure[$\xi = 0.2$]{
		\includegraphics[width = 0.24 \textwidth]{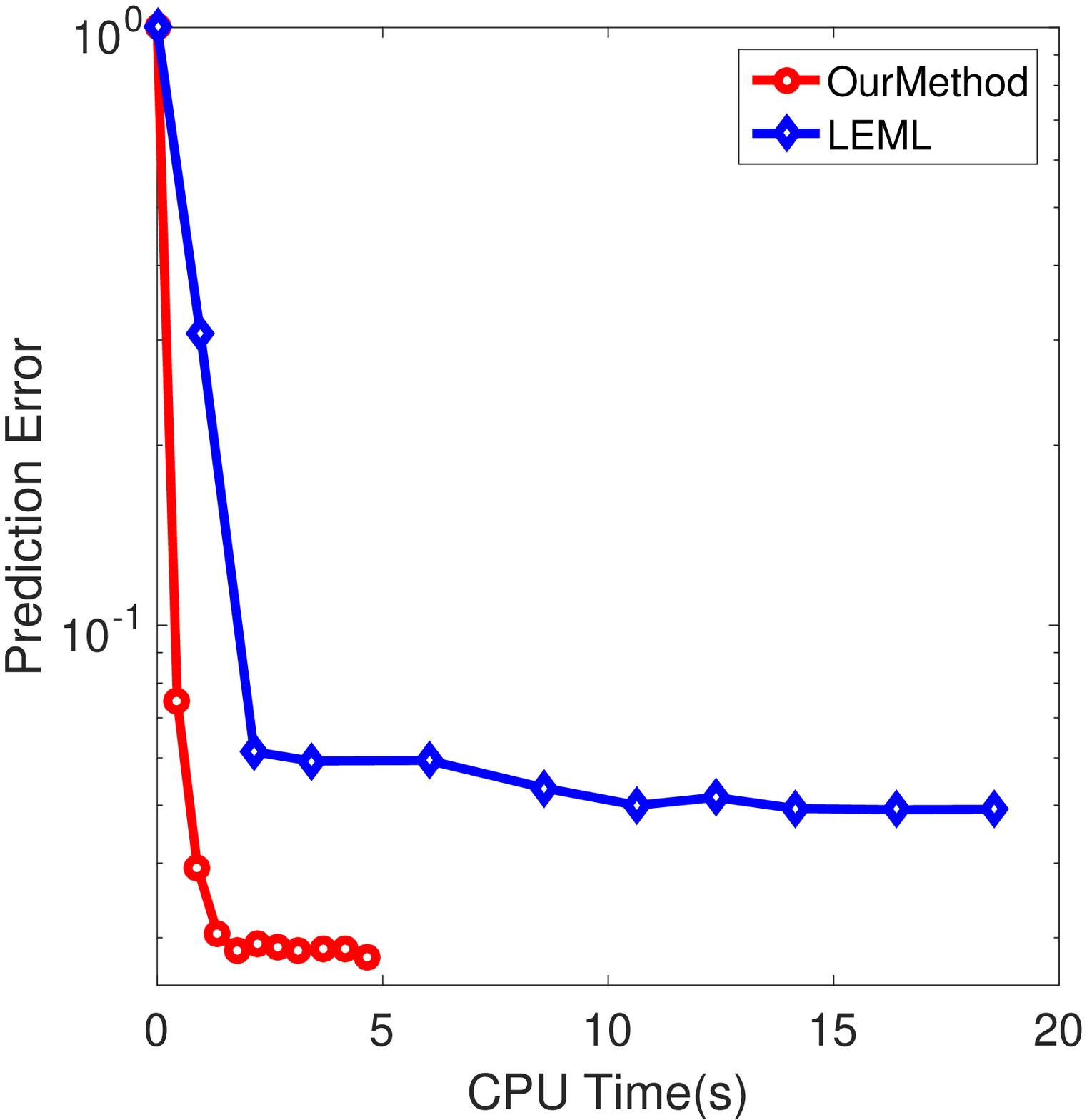}}
	\subfigure[$\xi = 0.3$]{
		\includegraphics[width = 0.24 \textwidth]{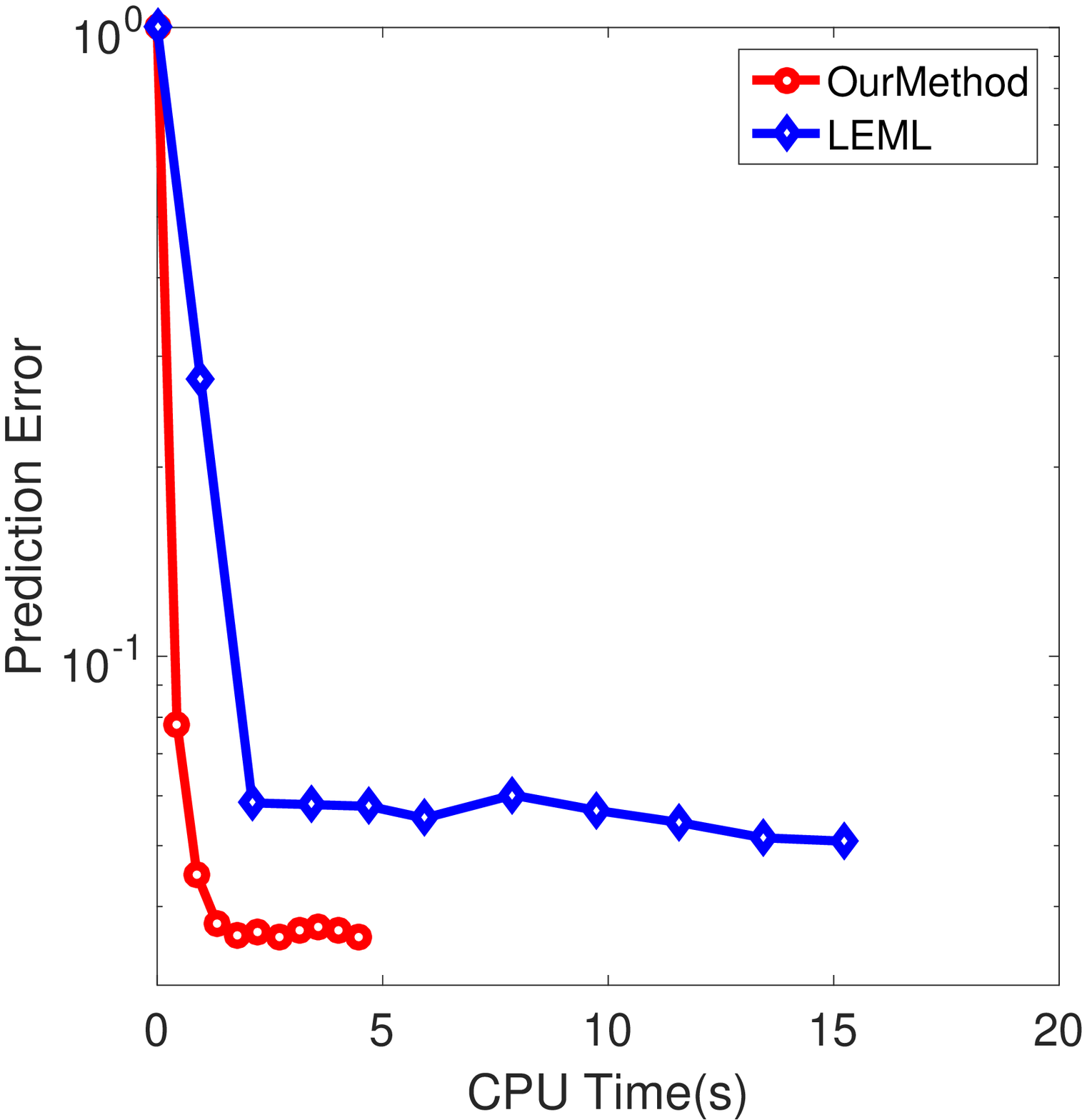}}
	\subfigure[$p = 1\%$]{
		\includegraphics[width = 0.24 \textwidth]{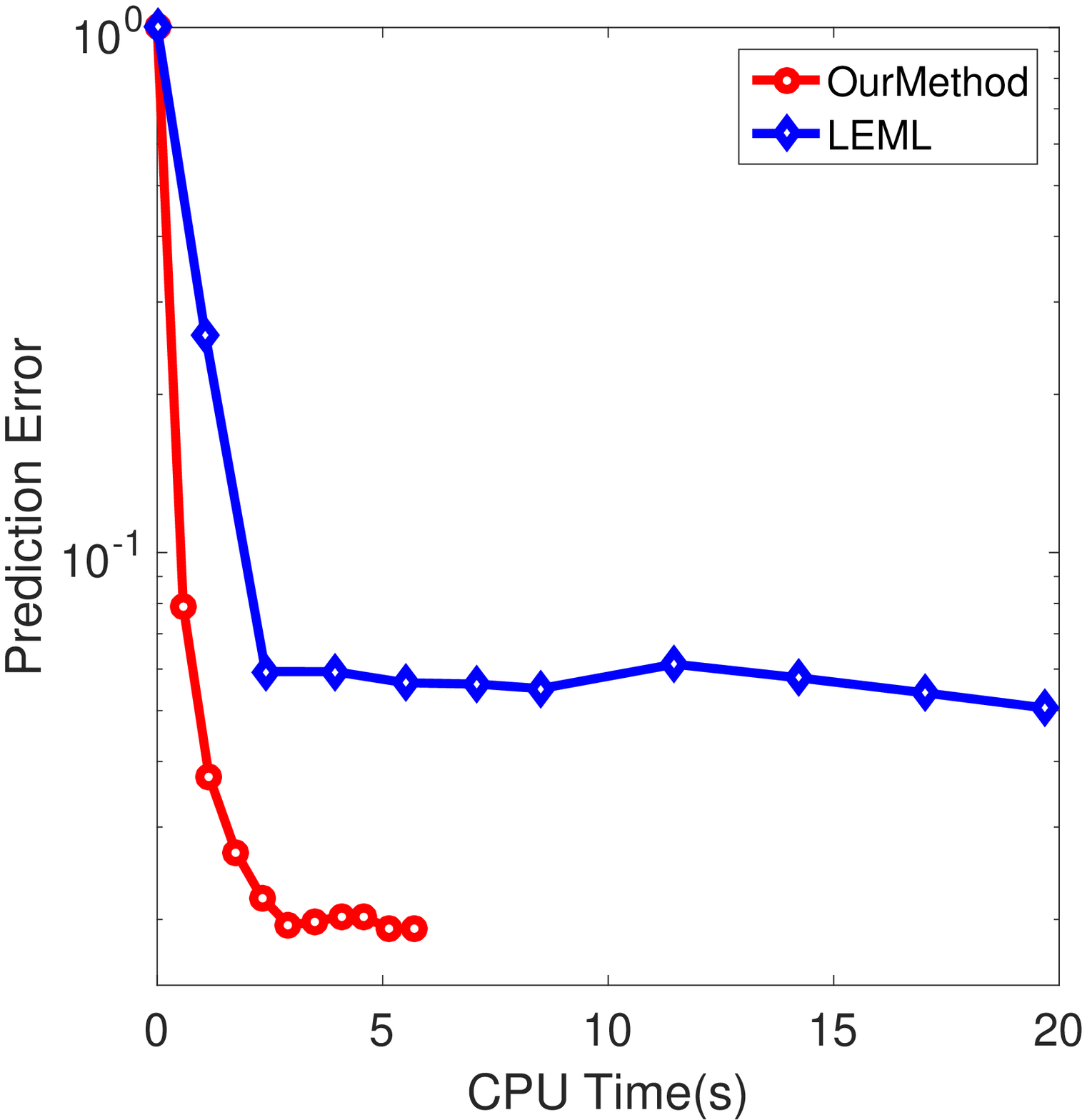}}
	\subfigure[$p = 2.5\%$]{
		\includegraphics[width = 0.24 \textwidth]{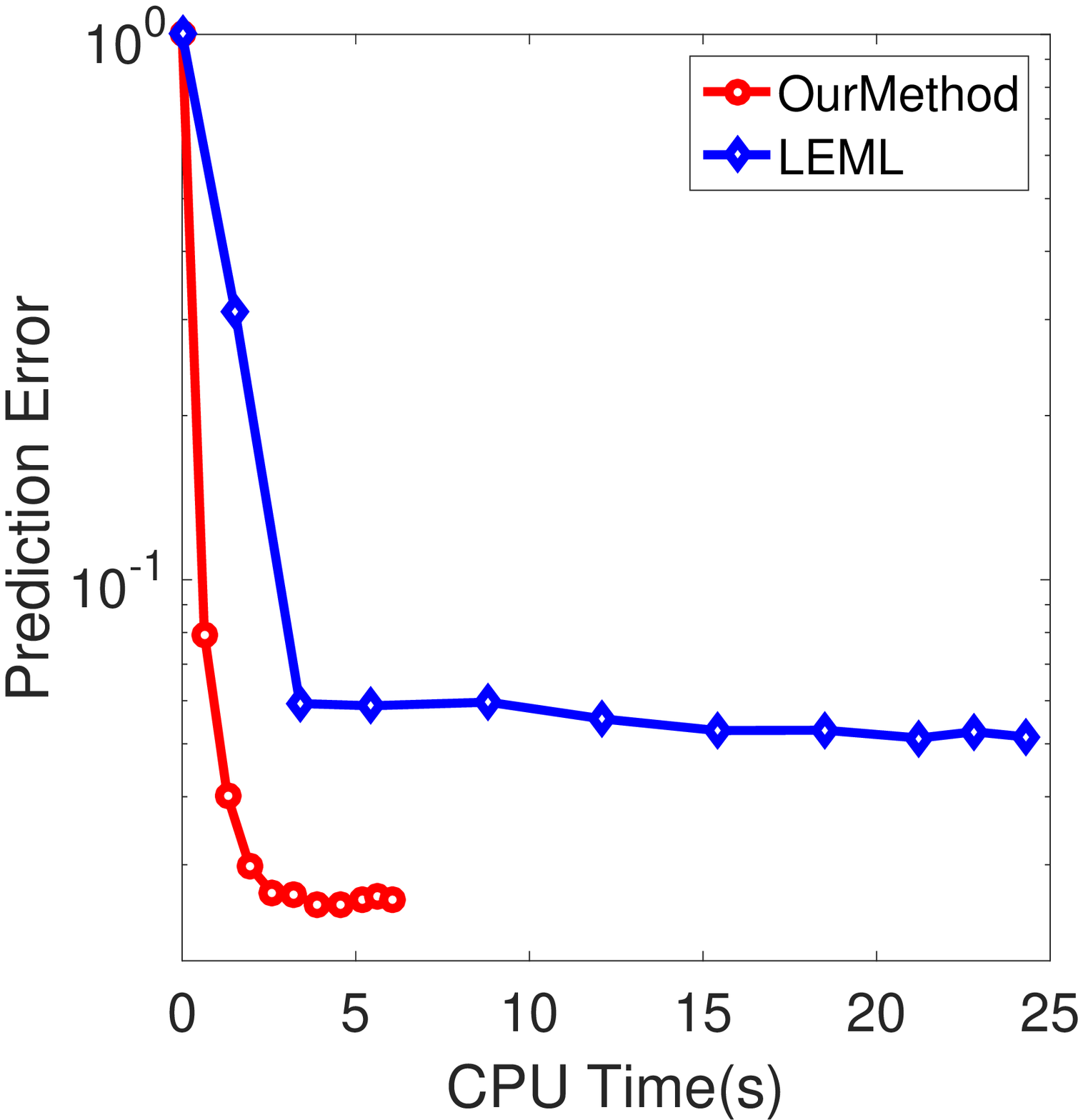}}
	\subfigure[$p = 5\%$]{
		\includegraphics[width = 0.24 \textwidth]{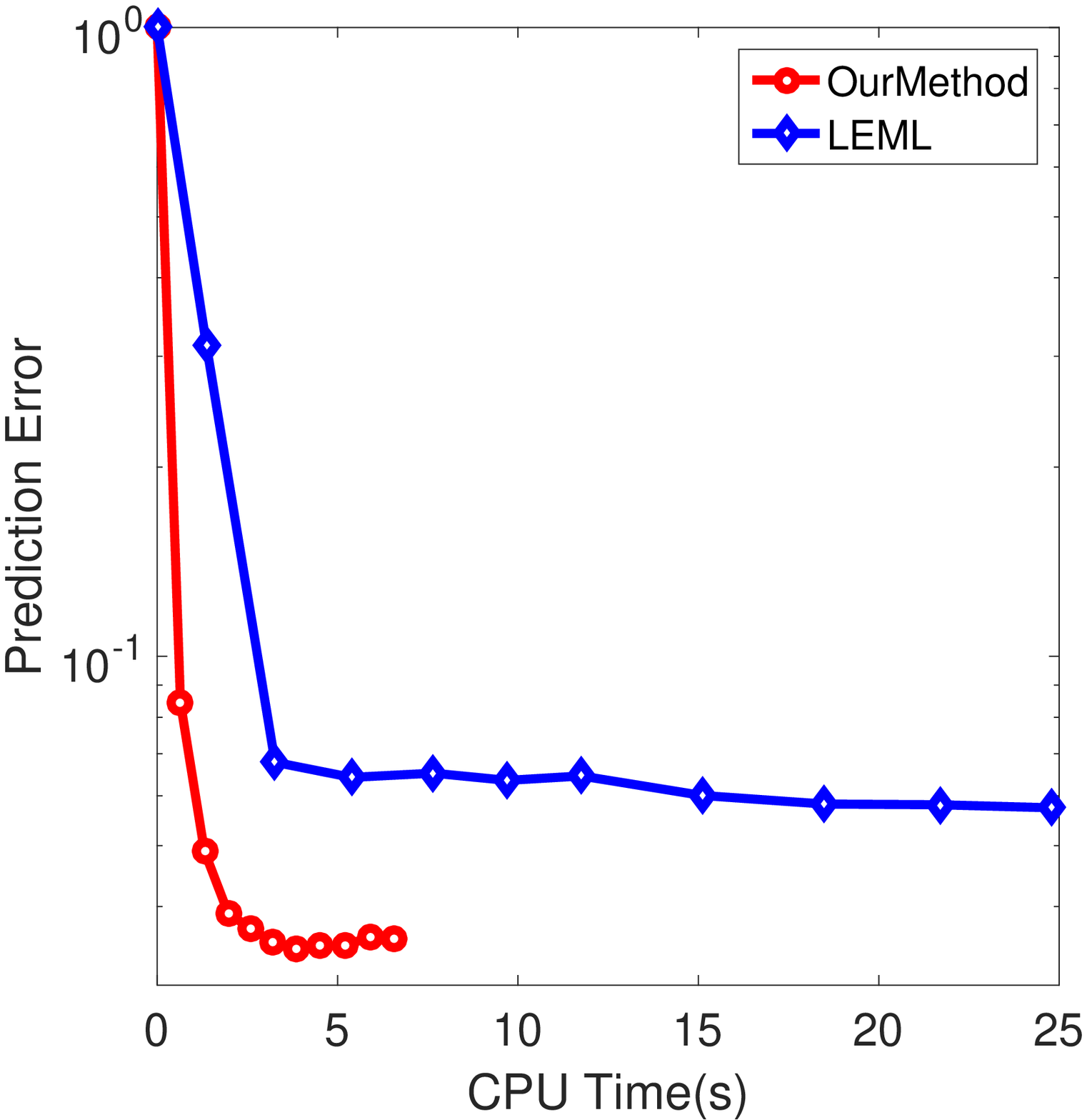}}
	\caption{Comparison results of our method and LEML on synthetic data.}\label{synthetic_results}
\end{figure*}

Our experimental results are shown in Fig.\ref{synthetic_results}, where we demonstrate the recover error and prediction error for the cases of no noise Fig.\ref{synthetic_results} (a)-(b), random noise Fig.\ref{synthetic_results} (c)-(e) and adversarial noise Fig.\ref{synthetic_results} (f)-(h). It can be observed that our algorithm achieves a lower prediction error on all dataset. Besides, our algorithm can converge globally with less convergence time compared with LEML. The average AUC is reported in Table \ref{table_noise_auc}, which also demonstrates that our method can outperform LEML in all the case . The results verifies the learning ability and the robustness to two distinct noises of our algorithm under this setting.

 \begin{table*}[!t]
	\renewcommand{\arraystretch}{1.1}
	\caption{\small The average AUC results.}
	\label{table_noise_auc}
	\centering
	\small{
		\begin{tabular}{|c|c|c|c|c|c|c|c|c|}
			\hline
			Method & Noise Free & $\xi = 0.1$ & $\xi = 0.2$ & $\xi = 0.3$ & $p = 1\%$ & $p = 2.5\%$ & $p = 5\%$ & $p = 10\%$ \\\hline
			LEML&95.29 &95.05 & 94.90 &94.30 & 95.02 &94.90 & 94.34 &93.42\\
			Ours&\textbf{98.73} & \textbf{97.90} & \textbf{97.19} &\textbf{96.52} & \textbf{98.66} &\textbf{97.47} & \textbf{96.64} & \textbf{95.79}\\
			\hline
	\end{tabular}}
\end{table*}
\subsection{Full-Observation Multi-label Learning}\label{multilabel}

Conventional multi-label learning problem settings sometimes assume that the training label matrix is fully observed. But we can extend our algorithm to this situation. During each iteration, our algorithm is able to randomly sample a batch of labels and the associated data instance to train the model.

In this subsection, we evaluate our method for the full-observation multi-label learning problem, which means the training data contains the whole labels for the training instances, which is the typical setting for the conventional multi-label learning problem. 

For our experiments, we compare our method with several multi-label methods: (1) multi-label SVM (SVM) without low rank constraint, (2) multi-label logistic regression with trace norm (LRT), (3) multi-label squared loss model with trace norm (SMT), and (4) LEML. We evaluate the effectiveness of our method on three synthetic datasets of different sizes. We set (1) $d_1 = 500$, $d_2 = 100$, $k = 3$ and $5000$ instances for Synthetic1, (2) $d_1 = 1000$, $d_2 = 300$, $k = 3$ and $10000$ instances for Synthetic2, and $d_1 = 2000$, $d_2 = 500$, $k = 3$ and $20000$ instances for Synthetic3. And then we test our model by a batch of $10000$ instances with their overall labels. Additionally, we compute the average AUC for each prediction result. The mean AUC results of all compared methods are presented in Table \ref{table_ml_auc}. The result in the table indicates that our algorithm can consistently outperform other methods, which testifies the ability of our algorithm on traditional multi-label problems.
 \begin{table*}[!t]
	\renewcommand{\arraystretch}{1.1}
	\caption{\small The average AUC results of different multi-label classification methods.}
	\label{table_ml_auc}
	\centering
	\begin{tabular}{|c|c|c|c|c|c|}
		\hline
		Data & SVM & LRT & SMT & LEML & Ours \\\hline
		Synthetic1&85.67 &93.23 &91.75  &90.55 & \textbf{93.94}\\
		Synthetic2&84.39 & 89.60  & 89.95 &91.38 & \textbf{95.70}\\
		Synthetic3&87.67 &93.70   &94.56 &95.18 & \textbf{98.15}\\
		\hline
	\end{tabular}
\end{table*}

%% file: lyx_conclusion.tex
\section{Conclusion}
In this paper, we investigate an extreme condition in multi-label learning where each training instance is generated with a single one-bit label out of multiple labels. We formulate this problem as a non-trivial special case of one-bit rank-one matrix sensing and develop an efficient non-convex algorithm based on alternating power iteration. Our algorithm can recover the underlying low-rank matrix model linearly. For a rank-$k$ model with $d_1$ features and $d_2$ classes, the proposed algorithm achieves $O(\epsilon)$ recovery error after retrieving $O(k^{1.5}d_1 d_2/\epsilon)$ one-bit labels within $O(kd)$ memory, which significantly improves the state-of-the-art sampling complexity of one-bit multi-label learning. We perform experiments to verify our theory and evaluate the performance of the proposed algorithm.

%% file: lyx_appendix.tex
\section{Preliminary}
\begin{theorem}[Matrix Bernstein's Inequalities~\cite{vershynin2016high, tropp2015introduction}]
Suppose $B_1,...B_m \in  \R^{d_1 \times d_2}$ are a set of independent random matrices of dimension $d_1 \times d_2$, and $$||B_i - \EE B_i||_2 \leq L, \forall i \in [m]$$ holds almost surely.
Define, 
$$
Z_i = B_i - \EE B_i, \quad \sigma^2 = \max \{||\sum_{i = 1}^{m} \EE(Z_i^\top Z_i) ||_2, ||\sum_{i =1 }^{m}\EE(Z_i Z_i^\top)||_2  \}
$$
Then, for all $t > 0$, 
$$
P(\frac{1}{m}||\sum_{i=1}^m Z_i||_2 \geq t) \leq (d_1 + d_2) \exp \bigg( \frac{-m^2 t^2}{ \sigma^2 + mLt /3 } \bigg)
$$
It can be further transformed as, with a probability at least $1-\eta$, the following can hold
$$
\frac{1}{m} ||\sum_{i = 1}^m Z_i ||_2 \leq \frac{4}{3} \frac{L}{m} \log(d_1 + d_2 /\eta) + 3 \sqrt{2\frac{\sigma^2}{m^2} \log((d_1 + d_2) /\eta)} 
$$
\end{theorem}

\section{Proof of Lemmas}

\subsection{Proof of Lemma~\ref{lemma_expectation}}
\begin{proof}
The paper~\cite{plan2013robust} shows a similar property in vector space. For the consistency of this paper, we rewrite it as follows.

For two normalized vector $\w',\w \in \R^{d_1}$, $|| \w||_2 = 1, || \w'||_2 = 1$, assuming $\boldsymbol g \in \R^{d_1}$ be the Gaussian vector with each entry following i.i.d. standard Gaussian distribution, we have
\begin{align*} 
&\EE\{\mathrm{sign}(\langle \boldsymbol g, \w \rangle) \langle \boldsymbol g, \w' \rangle\} \\
=& \EE\{\mathrm{sign}(\boldsymbol g) [ \boldsymbol g \langle \w, \w' \rangle + (1-\langle \w, \w' \rangle^2)^{1/2} \boldsymbol h]\}\\
=& \EE\{\mathrm{sign}(\boldsymbol g) \boldsymbol g\langle \w, \w' \rangle\} + \EE\{(1-\langle \w, \w' \rangle^2)^{1/2} \boldsymbol h\}\\
=& \EE\{|\boldsymbol g|\langle \w, \w' \rangle\} + 0\\
=& \sqrt{\frac{2}{\pi}}\langle \w, \w' \rangle
\end{align*}
where $\boldsymbol g$ and $\boldsymbol h$ are two independent Gaussian variables. 

Now we set $\w' = \e^{i}$ for any $i \in [d_1]$, where $\e^{i}$ is the matrix such that only the element at $i$ to be $1$ with all rest to be $0$. Note that $\e^i$ here is not a random vector. And we have
\begin{align*}
\EE\{\mathrm{sign}(\langle \boldsymbol g, \w \rangle) \langle \boldsymbol g, \e_{i} \rangle\} = \sqrt{\frac{2}{\pi}}\langle \w, \e_{i} \rangle 
\end{align*}
where implies that $\EE\{\mathrm{sign}(\langle \boldsymbol g, \w \rangle) \boldsymbol{g}_i = \sqrt{\frac{2}{\pi}} \w_{i}$. Thus we have
\begin{align*}
\EE\{\mathrm{sign}(\langle \boldsymbol g, \w \rangle) \boldsymbol g\} = \sqrt{\frac{2}{\pi}} \w
\end{align*}

We extend the above proof to our case. Assume each column of $W$ is normalized. Here we denote $\bar{\e}_i^j$ is the random vector, whose $j$-th position is $\sqrt{d_2}$ with $j$ following a Multinolli distribution. Then we have
\begin{align*}
&\EE\{\mathrm{sign}(\langle  \x_i \bar{\e}_i^\top, W \rangle)  \x_i \bar{\e}_i^\top\} \\
=& \EE_{\bar{\e}_i} \EE_{\x_i} \{\mathrm{sign}(\langle  \x_i ,  W \bar{\e}_i \rangle)  \x_i \bar{\e}_i^\top\}\\
=& \EE_{j} \EE_{\x_i} \{\mathrm{sign}(\langle  \x_i ,  W \bar{\e}_i^j \rangle)  \x_i \bar{\e}_i^j{}^\top\}\\
=& \EE_{j} \EE_{\x_i} \{\mathrm{sign}(\langle  \x_i ,  \sqrt{d_2} W_{\cdot, j} \rangle)  \x_i \bar{\e}_i^j{}^\top\}\\
=& \EE_{j} \EE_{\x_i} \{\mathrm{sign}(\langle  \x_i ,  W_{\cdot, j} \rangle)  \x_i \bar{\e}_i^j{}^\top\}\\
=& \sqrt{d_2} \EE_{j} \{\lambda  W_{\cdot, j} \bar{\e}_i^j{}^\top\}\\
=& \frac{\lambda}{\sqrt{d_2}} W
\end{align*}
which then further implies
\begin{align*}
\frac{1}{m} \EE\{\A' (\mathrm{sign}(\A(W)))\} = \frac{\lambda}{\sqrt{d_2}} W    
\end{align*}
\end{proof}

\subsection{Proof of Lemma~\ref{lemma_distance}}

Let $\alpha \triangleq \arccos(\langle \w, \w' \rangle)$ and $||\w||_2 = 1, ||\w'||_2 = 1$. We show the proof of the two inequalities in this lemma respectively under the condition that $\alpha \leq \frac{\pi}{2}$. Let $\boldsymbol g$ be a standard Gaussian vector $\boldsymbol g = [g_1, g_2,...,g_{d_1}]^\top$ with $g_i$ being the i.i.d. copy of $\mathcal{N}(0,1)$.

\textbf{(a)} Proof of $||\EE\{\boldsymbol g \boldsymbol g^\top |\mathrm{sign}(\langle \boldsymbol g , \w \rangle) - \mathrm{sign}(\langle \boldsymbol g , \w' \rangle) |^2 \}||_2 \leq C_1 ||\w - \w'||_2 $
\begin{proof}
We can define a rotation matrix $Q\in \R^{d_1 \times d_1}$ as follows, when $\w \neq \w'$ and $d_1 > 2$,
\begin{align*}
Q = [\w, \w' - \langle \w, \w'\rangle \w, M_\bot]
\end{align*}
Note that $M_\bot \in \R^{d_1 \times {d_1-2}}$ is a matrix whose each column is a set of orthonormal bases of the null space of $[\w,\w']$, such that $M_\bot^\top M_\bot = I$. It is easy to show that $Q^\top Q = I$ as well as $Q Q^\top  = I$ since $Q$ is full-rank.

Thus, we have
\begin{align*}
&||\EE\{Q^\top \boldsymbol g \boldsymbol g^\top Q |\mathrm{sign}(\langle  Q^\top \boldsymbol g , Q^\top \w \rangle) - \mathrm{sign}(\langle  Q^\top \boldsymbol g , Q^\top \w' \rangle) |^2 \}||_2 \\
=&||\EE\{\boldsymbol g \boldsymbol g^\top |\mathrm{sign}(\langle  Q^\top \boldsymbol g , Q^\top \w \rangle) - \mathrm{sign}(\langle  Q^\top \boldsymbol g , Q^\top \w' \rangle) |^2 \}||_2 \\
=&||\EE\{\boldsymbol g \boldsymbol g^\top |\mathrm{sign}(\langle \boldsymbol  g , \w \rangle) - \mathrm{sign}(\langle \boldsymbol g , \w' \rangle) |^2 \}||_2
\end{align*}
The last equality holds since $\langle  Q^\top \boldsymbol g , Q^\top \w \rangle = \langle   \boldsymbol g , \w \rangle$ and $\langle  Q^\top \boldsymbol g , Q^\top \w' \rangle = \langle \boldsymbol g , \w' \rangle$.

Due to the ratation invariance property of standard Gaussian vector, $Q^\top \boldsymbol g$ is equivalent to $
\boldsymbol g$ in distribution. And we also obtain that
$$
\bar{\w} \triangleq Q^\top \w = [1,0,...,0]^\top, \quad 
\bar{\w}' \triangleq Q^\top \w' = [\cos \alpha, \sin \alpha,0,...,0]^\top
$$
The above proof show that the term $||\EE\{\boldsymbol g \boldsymbol g^\top |\mathrm{sign}(\langle \boldsymbol g , \w \rangle) - \mathrm{sign}(\langle \boldsymbol g , \w' \rangle) |^2 \}||_2$ is rotation invariant to $\boldsymbol g$ and $\w,\w'$. And we design a ratation matrix $Q$ to simplify the computation of this term.

With the above formulation, we have
$$
||\EE\{\boldsymbol g \boldsymbol g^\top |\mathrm{sign}(\langle \boldsymbol  g , \w \rangle) - \mathrm{sign}(\langle \boldsymbol g , \w' \rangle) |^2 \}||_2 = ||\EE\{\boldsymbol g \boldsymbol g^\top |\mathrm{sign}(\langle \boldsymbol  g , \bar{\w} \rangle) - \mathrm{sign}(\langle \boldsymbol g , \bar{\w}' \rangle) |^2 \}||_2
$$

Now we rewrite it as 
\begin{align*}
&||\EE\{\boldsymbol g \boldsymbol g^\top |\mathrm{sign}(\langle \boldsymbol  g , \bar{\w} \rangle) - \mathrm{sign}(\langle \boldsymbol g , \bar{\w}' \rangle) |^2 \}||_2\\
=& \Bigg\| \EE \begin{bmatrix}
&g_1^2, &g_1 g_2,&....,&g_1 g_{d_1}\\
&g_2 g_1, &g_2^2,&...., &g_2 g_{d_1}\\
&...,&...,&...,&...\\
&g_{d_1} g_1, &g_{d_1} g_2,&....,&g_{d_1} g_{d_1}\\
\end{bmatrix} |\mathrm{sign}(\langle \boldsymbol  g , \bar{\w} \rangle) - \mathrm{sign}(\langle \boldsymbol g , \bar{\w}' \rangle) |^2  \Bigg\|
\end{align*}

Now we are going to compute each $\EE \{g_i g_j |\mathrm{sign}(\langle \boldsymbol  g , \bar{\w} \rangle) - \mathrm{sign}(\langle \boldsymbol g , \bar{\w}' \rangle) |^2 \}$. We can see that only when $g_1 > 0 \wedge g_1 \cos \alpha + g_2 \sin \alpha < 0$ or $g_1 < 0 \wedge g_1 \cos \alpha + g_2 \sin \alpha > 0$, the term $|\mathrm{sign}(\langle \boldsymbol  g , \bar{\w} \rangle) - \mathrm{sign}(\langle \boldsymbol g , \bar{\w}' \rangle) |^2 = 4$. Otherwise it is $0$. Therefore, the domain of integration for computing expectation is 
$$
\omega = \{(g_1,g_2) : g_1 > 0 \wedge g_1 \cos \alpha + g_2 \sin \alpha < 0 \} \cup \{(g_1,g_2) :  g_1 < 0 \wedge g_1 \cos \alpha + g_2 \sin \alpha > 0\}
$$
with all other Gaussian variables $g_3...g_{d_1} \in (-\infty, +\infty) $.

If $i = j = 1$, 
$$
\EE \{g_1^2 |\mathrm{sign}(\langle \boldsymbol  g , \bar{\w} \rangle) - \mathrm{sign}(\langle \boldsymbol g , \bar{\w}' \rangle) |^2 \} =\int_{(g_1,g_2) \in \omega} g_1^2 \phi(g_1)\phi(g_2)dg_1dg_2 \int_{g_3,...,g_{d_1}} \phi(g_3)...\phi(g_{d_1}) dg_3...dg_{d_1}  = 0
$$
And similarly, we can compute the other cases as follows.

If $i = j = 2$, 
\begin{align*}
&\EE \{g_2^2 |\mathrm{sign}(\langle \boldsymbol  g , \bar{\w} \rangle) - \mathrm{sign}(\langle \boldsymbol g , \bar{\w}' \rangle) |^2 \} \\
=&4\int_{(g_1,g_2) \in \omega} g_2^2 \phi(g_1)\phi(g_2)dg_1dg_2 \int_{g_3,...,g_{d_1}} \phi(g_3)...\phi(g_{d_1}) dg_3...dg_{d_1}\\
=&4\int_{(g_1,g_2) \in \omega} g_2^2 \phi(g_1)\phi(g_2)dg_1dg_2\\
=&8 \int_{\pi /2}^{\alpha + \pi /2} \int_{0}^{\infty} \sin^2(\theta) r^3 e^{-r^2} dr d\theta, \quad (substitution\ by \ polar\ coordinate\ for\ integration)\\
=& c_1 (2\alpha + \sin \alpha)
\end{align*}
since $\alpha \leq \frac{\pi}{2}$, the we have
$$
c_1(2+\frac{2}{\pi}) \alpha \leq c_1(2\alpha + \sin \alpha ) \leq 3c_1 \alpha
$$

If $i = j \geq 3$, we can have
$$
\EE \{g_i^2 |\mathrm{sign}(\langle \boldsymbol  g , \bar{\w} \rangle) - \mathrm{sign}(\langle \boldsymbol g , \bar{\w}' \rangle) |^2 \} = \frac{4\alpha}{\pi}
$$

If $i = 1, j = 2$ or $i = 2, j = 1$, 
\begin{align*}
&\EE \{g_1 g_2 |\mathrm{sign}(\langle \boldsymbol  g , \bar{\w} \rangle) - \mathrm{sign}(\langle \boldsymbol g , \bar{\w}' \rangle) |^2 \} \\
=&4\int_{(g_1,g_2) \in \omega} g_1 g_2 \phi(g_1)\phi(g_2)dg_1dg_2 \int_{g_3,...,g_{d_1}} \phi(g_3)...\phi(g_{d_1}) dg_3...dg_{d_1}\\
=&4\int_{(g_1,g_2) \in \omega} g_1 g_2 \phi(g_1)\phi(g_2)dg_1dg_2\\
=&4 \int_{\pi /2}^{\alpha + \pi /2} \int_{0}^{\infty} \sin(2\theta) r^3 e^{-r^2} drd\theta, \quad (substitution\ by \ polar\ coordinate\ for\ integration)\\
=& - c_2 \sin^2 \alpha
\end{align*}
since $\alpha \leq \frac{\pi}{2}$, the we have
$$
- c_2 \alpha^2 \leq - c_2 \sin^2 \alpha \leq  -c_2 \frac{4}{\pi^2}\alpha^2
$$

For all the other case that $i\neq j$, we can get that 
\begin{align*}
&\EE \{g_i g_j |\mathrm{sign}(\langle \boldsymbol  g , \bar{\w} \rangle) - \mathrm{sign}(\langle \boldsymbol g , \bar{\w}' \rangle) |^2 \} = 0
\end{align*}

Thus, we have
\begin{align*}
&||\EE\{\boldsymbol g \boldsymbol g^\top |\mathrm{sign}(\langle \boldsymbol  g , \bar{\w} \rangle) - \mathrm{sign}(\langle \boldsymbol g , \bar{\w}' \rangle) |^2 \}||_2\\
=& \Bigg\| \begin{bmatrix}
&0, &- c_2 \sin^2 \alpha ,&0,&....,&0\\
&- c_2 \sin^2 \alpha , &c_1 (2\alpha + \sin \alpha),&0,&...., &0\\
&0,&0,&4\alpha / \pi ,&...,&0\\
&...,&...,&...,&...,&...\\
&0,&0, &0, &....,&4\alpha / \pi\\
\end{bmatrix} \Bigg\|\\
= &\Bigg\| \begin{bmatrix}
&0, &- c_2 \sin^2 \alpha / \alpha,&0,&....,&0\\
&- c_2 \sin^2 \alpha / \alpha, &c_1 (2 + \sin \alpha / \alpha),&0,&...., &0\\
&0,&0,&4 / \pi ,&...,&0\\
&...,&...,&...,&...,&...\\
&0,&0, &0, &....,&4 / \pi\\
\end{bmatrix} \Bigg\| \alpha  \\
\leq & C' \alpha \\
\leq &C_1 ||\w-\w'||_2
\end{align*}
\end{proof}
since all the terms in the above matrix can be bounded after being divided by $\alpha$. The last inequality can be easily proved if $\alpha \leq \frac{\pi}{2}$. We omit the proof here.

\textbf{(b)} Proof of $\EE\{ ||\boldsymbol g||_2^2  |\mathrm{sign}(\langle \boldsymbol g , \w \rangle) - \mathrm{sign}(\langle \boldsymbol g , \w' \rangle) |^2 \} \leq C_2 d_1 ||\w - \w'||_2 $

\begin{proof}
We can apply the same technique as the above proof. And we have that
\begin{align*}
&\EE\{ ||\boldsymbol g||_2^2  |\mathrm{sign}(\langle \boldsymbol g , \w \rangle) - \mathrm{\mathrm{sign}}(\langle \boldsymbol g , \w' \rangle) |^2 \}\\
=& \EE\{||\boldsymbol g ||_2^2 |\mathrm{sign}(\langle \boldsymbol  g , \bar{\w} \rangle) - \mathrm{\mathrm{sign}}(\langle \boldsymbol g , \bar{\w}' \rangle) |^2 \}\\
= & \sum_{i} g_i^2 |\mathrm{sign}(\langle \boldsymbol  g , \bar{\w} \rangle) - \mathrm{sign}(\langle \boldsymbol g , \bar{\w}' \rangle) |^2 \\
\leq & 0 + 3c_1 \alpha + (d_1-2) \frac{4\alpha}{\pi}  \\
\leq & C'' d_1 \alpha \\
\leq & C_2 d_1 ||\w-\w' ||_2
\end{align*}
\end{proof}

\subsection{Proof of Lemma~\ref{lemma_loss}}
Remember in Algorithm 1, we $\W^{(t)}$ is constructed by first extracting $\widetilde{W}^{(t)}$ from $\widetilde{W}^{(t)}$ and then using Hermitian Dilation to the column normalized $W^{(t)}$.Note that in the below proof, we use the property that $||W^*_{\cdot,j}||_2 = ||W^{(t)}_{\cdot,j}||_2 = 1$ for all $j \in [d_2]$.

We will analyze the column normalization first. Suppose the column vector $W^{(t)}_{\cdot, j}$ is the normalized vector of $\widetilde{W}^{(t)}_{\cdot, j}$
\begin{align*}
&||W^{(t)}_{\cdot, j}-\widetilde{W}^{(t)}_{\cdot, j}||_2 \\
= & \sqrt{||\widetilde{W}^{(t)}_{\cdot, j}||_2^2 + ||W^{(t)}_{\cdot, j}||_2^2 - 2 \langle \widetilde{W}^{(t)}_{\cdot, j}, W^{(t)}_{\cdot, j} \rangle} \\
= & \sqrt{||\widetilde{W}^{(t)}_{\cdot, j}||_2^2 + ||W^*_{\cdot, j}||_2^2 - 2 \langle \widetilde{W}^{(t)}_{\cdot, j}, W^{(t)}_{\cdot, j} \rangle} \\
\leq & \sqrt{||\widetilde{W}^{(t)}_{\cdot, j}||_2^2 + ||W^*_{\cdot, j}||_2^2 - 2 \langle \widetilde{W}^{(t)}_{\cdot, j}, W^*_{\cdot, j} \rangle}\\
= & ||W^*_{\cdot, j}-\widetilde{W}^{(t)}_{\cdot, j}||_2 
\end{align*}
The last but one inequality in the above formulation is due to $\langle \widetilde{W}^{(t)}_{\cdot, j}, W^{(t)}_{\cdot, j} \rangle \geq \langle \widetilde{W}^{(t)}_{\cdot, j}, W^*_{\cdot, j} \rangle$ because the angle between $\widetilde{W}^{(t)}_{\cdot, j}, W^{(t)}_{\cdot, j}$ is 0. It is easy to understand the above formulation since $W^{(t)}_{\cdot, j}$ , rather than $W^*_{\cdot, j}$, is the nearest point to $\widetilde{W}^{(t)}_{\cdot, j}$ on a unit sphere .

The above further implies that
\begin{align*}
||W^{(t)}-\widetilde{W}^{(t)}||_F \leq ||W^*-\widetilde{W}^{(t)}||_F
\end{align*}
if we take the sum of the above inequality through all the columns.

We can extend the above inequality to $\widetilde{\W}^{(t)}$ as follows,
\begin{align*}
&||\mathcal{W}^{(t)}-\widetilde{\mathcal{W}}^{(t)}||_F \\
=& \Bigg \|\begin{bmatrix}
O & W^{(t)}\\ 
W^{(t)}{}^\top  & O
\end{bmatrix} -\begin{bmatrix}
N_1 & \widetilde{W}^{(t)}\\ 
\widetilde{W}^{(t)}{}^\top  & N_2
\end{bmatrix}\Bigg \|_F \\
=& \sqrt{||N_1||^2_F+||N_2||^2_F+2||W^{(t)}-\widetilde{W}^{(t)}||^2_F}\\
\leq& \sqrt{||N_1||^2_F+||N_2||^2_F+2||W^*-\widetilde{W}^{(t)}||^2_F}\\
= & \Bigg\|\begin{bmatrix}
O & W^* \\ 
{W^*}^\top  & O
\end{bmatrix} -\begin{bmatrix}
N_1 & \widetilde{W}^{(t)} \\ 
\widetilde{W}^{(t)}{}^\top & N_2
\end{bmatrix}\Bigg\|_F \\
= & ||\mathcal{W}^* - \mathcal{\widetilde{W}}^{(t)}||_F 
\end{align*}
where $N_1$ and $N_2$ are other block submatrices in the matrix $\widetilde{\W}^{(t)}$.

And thus we can have
\begin{align*}
&||\mathcal{W}^*-\mathcal{W}^{(t)}||_2 \\
\leq & ||\mathcal{W}^*-\mathcal{W}^{(t)}||_F \\
\leq&||\mathcal{W}^*-\mathcal{\widetilde{W}}^{(t)}||_F + ||\mathcal{W}^{(t)}-\mathcal{\widetilde{W}}^{(t)}||_F \\
\leq&2||\mathcal{W}^*-\mathcal{\widetilde{W}}^{(t)}||_F\\
\leq&4\sqrt{k}||\mathcal{W}^*-\mathcal{\widetilde{W}}^{(t)}||_2
\end{align*}
since $rank(\mathcal{W}^*) = rank(\mathcal{\widetilde{W}}^{(t)}) \leq 2k$ and then the rank of $\mathcal{W}^*-\mathcal{\widetilde{W}}^{(t)}$ is less than or equal to $4k$. 

\subsection{Proof of Lemma~\ref{lemma_angle}}
\begin{proof} It is not difficult to show our proof as that
$\tan \theta(\U^*, \widetilde{\U}^{(t)})  =||\U^*_{\bot}{}^\top \widetilde{\U}^{(t)} (\U^*{}^\top \widetilde{\U}^{(t)})^{-1} ||_2 
= ||\U^*_{\bot}{}^\top {\U}^{(t)}R (\U^*{}^\top {\U}^{(t)} R)^{-1} ||_2 
= ||\U^*_{\bot}{}^\top {\U}^{(t)}(\U^*{}^\top {\U}^{(t)})^{-1} ||_2  = \tan \theta(\U^*, {\U}^{(t)})$, which completes the proof.
\end{proof}

\subsection{Proof of Lemma~\ref{lemma_descent}}
Under the condition in this lemma, we split the proof into two cases, $\epsilon_t > 1$ and $\epsilon_t \leq 1$.

\textbf{(a)} $\epsilon_t > 1$
\begin{proof}
For $\alpha_{t+1}$, we can derive
\begin{align*}
&||\U^*_\bot (\W^*+ O(\delta \epsilon_t)) \U^{(t)}||_2 \\
\leq& ||\U^*_\bot W^* \U^*_\bot||_2 + \delta \epsilon_{t}\\
\leq& ||\U^*_\bot \U^* \Lambda^* \U^*{}^\top \U^*_\bot||_2 + \delta \epsilon_{t}\\
\leq& 0 + \delta \epsilon_{t}\\
\leq& \delta \epsilon_{t}
\end{align*}
\begin{align*}
&\sigma_{2k}\{\U^* (\W^*+ O(\delta \epsilon_t))\U^{(t)}\} \\
\geq &\U^*{}^\top \W^* \U^{(t)} - \delta \epsilon_t\\
\geq & \sigma_{2k}(\W^*) \sigma_{2k}\{ \U^*{}^\top \U^{(t)} \} -\delta \epsilon_t\\
=&\sigma_{k}^*\sigma_{2k}\{ \U^*{}^\top \U^{(t)} \} -\delta \epsilon_t\\
=&\sigma_{k}^* \cos\theta_t -\delta \epsilon_t
\end{align*}
Therefore, we can obtain
\begin{align*}
\alpha_{t+1} =& \frac{||\U^*_\bot (\W^*+ O(\delta \epsilon_t)) \U^{(t)}||_2}{\sigma_{2k}\{\U^* (\W^*+ O(\delta \epsilon_t))\U^{(t)}\}}\\
\leq & \frac{\delta\epsilon_t}{\sigma^*_k \cos \theta_t - \delta \epsilon_t}
\end{align*}
According to the assumption, $\cos \theta_t \geq \frac{1}{\sqrt{5}}$, $\delta \epsilon_t \leq \frac{2}{3\sqrt{5}} \sigma^*_k$, we thus have
\begin{align*}
\alpha_{t+1} \leq 3\sqrt{5}\delta \epsilon_t / \sigma^*_k
\end{align*}

For $\epsilon_{t+1}$, by lemma~\ref{lemma_loss}, we can have that
\begin{align*}
    \epsilon_{t+1} &= ||\W^* - \W^{(t+1)} ||_2 \\
    &\leq 4\sqrt{k}||\W^* - \widetilde{\W}^{(t+1)} ||_2,\quad \text{(by lemma~\ref{lemma_loss})}\\
    &=4\sqrt{k}||\W^* - (\U^{(t+1)} \U^{(t+1)}{}^\top (\H^{(t)} + \W^{(t)})^\top) ||_2\\
    &=4\sqrt{k}||\W^*-\U^{(t+1)}\U^{(t+1)}{}^\top (\W^* + O(\delta\epsilon_t)^\top)||_2\\
    &=4\sqrt{k}||(I - \U^{(t+1)}\U^{(t+1)}{}^\top)\W^* + \U^{(t+1)}\U^{(t+1)}{}^\top O(\delta\epsilon_t)^\top||_2\\
    &\leq 4\sqrt{k}||(I - \U^{(t+1)}\U^{(t+1)}{}^\top)\W^* ||_2 + 4\sqrt{k}||\U^{(t+1)}\U^{(t+1)}{}^\top O(\delta\epsilon_t)^\top||_2\\
    &\leq 4\sqrt{k}||( I-\mathcal{U}^{(t+1)}(\mathcal{U}^{(t+1)}{}^\top)\mathcal{U}^* \Lambda^{*}\mathcal{U}^*{}^\top||_2 + 4\sqrt{k}\delta\epsilon_t\\
    &\leq 4\sqrt{k}||( I-\mathcal{U}^{(t+1)}(\mathcal{U}^{(t+1)}){}^\top)\mathcal{U}^*||_2 \cdot || \Lambda^{*}\mathcal{U}^*{}^\top||_2 + 4\sqrt{k}\delta \epsilon_t\\
    &=4\sqrt{k}\sin \theta_{t+1} || \Lambda^{*}(\mathcal{U}^*){}^\top||_2 + 4\sqrt{k}\delta\epsilon_t\\
    &\leq 4\sqrt{k} \tan \theta_{t+1}||\W^*||_2 + 4\sqrt{k} \delta\epsilon_t\\
    &= 4\sqrt{k} \tan \theta_{t+1}\sigma^*_1 + 4\sqrt{k} \delta\epsilon_t\\
\end{align*}

\end{proof}

\textbf{(b)} $\epsilon_t \leq 1$
The proof is similar to (a), but there is also a little difference. The $\epsilon_t$ in (a) becomes $\epsilon_t^{1/2}$ here.
\begin{proof}

For $\alpha_{t+1}$, we can derive
\begin{align*}
&||\U^*_\bot (\W^*+ O(\delta \epsilon_t^{1/2})) \U^{(t)}||_2 \\
\leq& ||\U^*_\bot W^* \U^*_\bot||_2 + \delta \epsilon_{t}^{1/2}\\
\leq& \delta \epsilon_{t}^{1/2}
\end{align*}
\begin{align*}
&\sigma_{2k}\{\U^* (\W^*+ O(\delta \epsilon_t^{1/2}))\U^{(t)}\} \\
\geq & \sigma_{2k}(\W^*) \sigma_{2k}\{ \U^*{}^\top \U^{(t)} \} -\delta \epsilon_t^{1/2}\\
=&\sigma_{k}^* \cos\theta_t -\delta \epsilon_t^{1/2}
\end{align*}
Therefore, we can obtain
\begin{align*}
\alpha_{t+1} \leq \frac{\delta\epsilon_t^{1/2}}{\sigma^*_k \cos \theta_t - \delta \epsilon_t^{1/2}}
\end{align*}
According to the assumption, $\cos \theta_t \geq \frac{1}{\sqrt{5}}$, $\delta \epsilon_t^{1/2} \leq \frac{2}{3\sqrt{5}} \sigma^*_k$, we thus have
\begin{align*}
\alpha_{t+1} \leq 3\sqrt{5}\delta \epsilon_t^{1/2} / \sigma^*_k
\end{align*}

For $\epsilon_{t+1}$, similarly, we can have that
\begin{align*}
    \epsilon_{t+1} &= ||\W^* - \W^{(t+1)} ||_2 \\
    &\leq 4\sqrt{k}||\W^* - \widetilde{\W}^{(t+1)} ||_2,\quad \text{(by lemma~\ref{lemma_loss})}\\
    &=4\sqrt{k}||\W^* - (\U^{(t+1)} \U^{(t+1)}{}^\top (\H^{(t)} + \W^{(t)})^\top) ||_2\\
    &=4\sqrt{k}||\W^*-\U^{(t+1)}\U^{(t+1)}{}^\top (\W^* + O(\delta\epsilon_t^{1/2})^\top)||_2\\
    &\leq4\sqrt{k}\sin \theta_{t+1} || \Lambda^{*}(\mathcal{U}^*){}^\top||_2 + 4\sqrt{k}\delta\epsilon_t^{1/2}\\
    &\leq 4\sqrt{k} \tan \theta_{t+1}||\W^*||_2 + 4\sqrt{k} \delta\epsilon_t^{1/2}\\
    &= 4\sqrt{k} \tan \theta_{t+1}\sigma^*_1 + 4\sqrt{k} \delta\epsilon_t^{1/2}\\
\end{align*}

\end{proof}

\section{Proof of Theorems}

\subsection{Proof of Theorem~\ref{theorem_diffrip}}

\begin{proof}    
We set 
\begin{align*}
B_{i} = \frac{\sqrt{d_2}}{\lambda} [\x_i \bar{\e}_i^\top \mathrm{sign}(\langle \x_i \bar{\e}_i^\top, W \rangle) - \x_i \bar{\e}_i^\top \mathrm{sign}(\langle \x_i \bar{\e}_i^\top, W' \rangle)]
\end{align*}
And by lemma~\ref{lemma_expectation}, we have
\begin{align*}
\EE B_{i} = W-W' 
\end{align*}
We further let
$$Z_{i} = \sum_{i=1}^m B_{i} - \EE B_{i}$$
which implies
$$
\sum_{i = 1}^{m} B_{i} = \frac{\sqrt{d_2}}{\lambda} [\A' \mathrm{sign}(\A(W)) - \A' \mathrm{sign}(\A(W'))] 
$$

In addition, since $\x_i$ is an unbounded variable, we provide an high probability bound for $||\x_i||_2$ which is used by the following proofs. For all $(i,j) \in \Omega$, with a probability as least $1-\eta$, if $d_1 \geq  8\log(2n/\eta)$, then we have
\begin{align}\label{gaussian_bound}
 ||\x_i||_2^2 \leq 2d_1, \forall  i \in [m] 
\end{align}

With the above settings, we can bound the terms $\max_{i} ||Z_{i}||_2$, $||\EE\{Z_{i}^\top Z_{i}\}||_2$ and $||\EE\{Z_{i} Z_{i}^\top\}||_2$ such that we are able to utilize matrix Bernstein's inequality to derive our result.

\noindent \textbf{Bound $\max_{i} ||Z_{i}||_2$}
\begin{align*}
\quad &\max_{i} ||Z_{i}||_2 \\
=&\max_{i} ||(B_{i} - \EE B_{i})||_2 \\
\leq& \max_{i} (||B_{i}||_2 + ||\EE B_{i}||_2) \\
\leq& \max_{i} \frac{\sqrt{d_2}}{\lambda}||\x_i \bar{\e}_i^\top [\mathrm{sign}(\langle \x_i \bar{\e}_i^\top, W \rangle) - \mathrm{sign}(\langle \x_i \bar{\e}_i^\top, W' \rangle)]||_2 + || W - W'||_2 \\
\leq& \max_{i} (\frac{2\sqrt{d_2}}{\lambda}||\x_i \bar{\e}_i^\top||_2 + || W - W'||_2) 
\end{align*}
in which the last inequality holds because $|\mathrm{sign}(\langle \x_i \bar{\e}_i^\top, W \rangle) - \mathrm{sign}(\langle \x_i \bar{\e}_i^\top, W' \rangle)| \leq 2$. 

Since $||\x_i \bar{\e}_i^\top||_2 = \sqrt{d_2}||\x_i||_2$, by Formula~\ref{gaussian_bound}, with a probability at least $1-\eta_1$, if $d \geq  8\log(2m/\eta_1)$, then there is  
\begin{align*}
\max_{i} ||\x_i \bar{\e}_i^\top||_2 = \max_{i} \sqrt{d_2}||\x_i||_2\leq \sqrt{d_2} \sqrt{2d_1} = \sqrt{2d_1d_2}  
\end{align*}

The above implies that
\begin{align*}
\quad &\max_{i} ||Z_{i}||_2 \\
=&\max_{i} ||(B_{i} - \EE B_{i})||_2 \\
\leq& \max_{i} (||B_{i}||_2 + ||\EE B_{i}||_2) \\
\leq& \max_{i} (\frac{2\sqrt{d_2}}{\lambda}||\x_i \bar{\e}_i^\top||_2 + || W - W'||_2) \\
\leq& \frac{2\sqrt{2d_1} d_2}{\lambda} + || W - W'||_2
\end{align*}
with a probability at least $1-\eta_1$ if $d_1 \geq 8\log(2n/\eta_1)$. And if $||W-W'||_2$ is sufficiently small, 
\begin{align*}
\frac{2\sqrt{2d_1} d_2}{\lambda} + || W - W'||_2 \leq c \frac{2\sqrt{2d_1} d_2}{\lambda}
\end{align*}
And it is also easy to show with the below proof that this can hold.

\noindent \textbf{Bound $||\sum_{i=1}^m \EE\{ Z_{i}^\top Z_{i} \}||_2$}

Since $||\sum_{i=1}^m \EE\{ Z_{i}^\top Z_{i} \}||_2 \leq \sum_{i=1}^m ||\EE\{ Z_{i}^\top Z_{i} \}||_2$, we just need to bound $||\EE\{ Z_{i}^\top Z_{i} \}||_2$.

\begin{align*}
&||\EE\{ Z_{i}^\top Z_{i} \}||_2\\
=&||\EE\{ (B_{i}-\EE B_{i})^\top (B_{i}-\EE B_{i})   \}||_2\\
=&  ||\EE\{ B_{i}^\top B_{i} -\EE B_{i}^\top \cdot B_{i} - B_{i}^\top \cdot \EE B_{i} + \EE B_{i}^\top \EE B_{i}\}||_2\\
=&  ||\EE\{ B_{i}^\top B_{i} \} - \EE B_{i}^\top \EE B_{i}||_2\\
\leq& ||\EE\{ B_{i}^\top B_{i} \}||_2 + ||\EE B_{i}^\top \EE B_{i}||_2
\end{align*}

We bound the terms $||\EE\{ B_{i}^\top B_{i} \}||_2$ and $||\EE B_{i}^\top \EE B_{i}||_2$ respectively.
$$
||\EE B_{i}^\top \EE B_{i}||_2 = ||W-W'||^2_2
$$

\begin{align*}
&||\EE\{ B_{i}^\top B_{i} \}||_2\\
=&\frac{d_2}{\lambda^2}||\EE\{  (\x_i \bar{\e}_i^\top)^\top\x_i \bar{\e}_i^\top |\mathrm{sign}(\langle \x_i \bar{\e}_i^\top, W \rangle) -\mathrm{sign}(\langle \x_i \bar{\e}_i^\top, W' \rangle)|^2 \}||_2 \\
=& \frac{d_2}{\lambda^2}||\EE\{  || \x_i||_2^2 \bar{\e}_i \bar{\e}_i^\top |\mathrm{sign}(\langle \x_i \bar{\e}_i^\top, W \rangle) -\mathrm{sign}(\langle \x_i \bar{\e}_i^\top, W' \rangle)|^2 \}||_2 \\
=& \frac{d_2}{\lambda^2}|| \EE_{\x_i} \EE_{\bar{\e}_i}\{  || \x_i||_2^2 \bar{\e}_i \bar{\e}_i^\top |\mathrm{sign}(\langle \x_i \bar{\e}_i^\top, W \rangle) -\mathrm{sign}(\langle \x_i \bar{\e}_i^\top, W' \rangle)|^2 \}||_2 
\end{align*}

We denote $\bar{\e}_i^j{}^\top$ as a random vector whose $j$-th element is $\sqrt{d_2}$ with the $j$ follows the multinoulli distribution taking values from $1,...,d_2$ with equal probability $\frac{1}{d_2}$. And by the lemma~\ref{lemma_distance}, we can have

\begin{align*}
&\frac{d_2}{\lambda^2}|| \EE_{\x_i} \EE_{\bar{\e}_i}\{  || \x_i||_2^2 \bar{\e}_i \bar{\e}_i^\top |\mathrm{sign}(\langle \x_i \bar{\e}_i^\top, W \rangle) -\mathrm{sign}(\langle \x_i \bar{\e}_i^\top, W' \rangle)|^2 \}||_2 \\
=&\frac{d_2}{\lambda^2}|| \EE_{\x_i} \EE_{j}\{  || \x_i||_2^2 \bar{\e}_i^j \bar{\e}_i^j{}^\top |\mathrm{sign}(\langle \x_i , W \bar{\e}_i^j{}^\top \rangle) -\mathrm{sign}(\langle \x_i , W' \bar{\e}_i^j{}^\top \rangle)|^2 \}||_2 \\
\leq&\frac{d_2 }{\lambda^2}  \Bigg\|\frac{1}{d_2} \begin{bmatrix}
&d_2 h_1 ,  &0, &..., &0\\
&0, &d_2 h_2, &..., &0\\
&..., &..., &..., &...\\
&0, &0, &..., & d_2 h_{d_2}
\end{bmatrix}
\Bigg \|_2 \\
=& \frac{d_2 }{\lambda^2}  \Bigg\| \begin{bmatrix}
&h_1 ,  &0, &..., &0\\
&0, &h_2, &..., &0\\
&..., &..., &..., &...\\
&0, &0, &..., & h_{d_2}
\end{bmatrix}
\Bigg \|_2 \\
\leq& \frac{d_2 }{\lambda^2}  \Bigg\| \begin{bmatrix}
&h_1 ,  &0, &..., &0\\
&0, &h_2, &..., &0\\
&..., &..., &..., &...\\
&0, &0, &..., & h_{d_2}
\end{bmatrix}
\Bigg \|_F \\
=& \frac{ d_2 }{\lambda^2} \sqrt{\sum_{j = 1}^{d_2} h_j^2} \\
\leq& \frac{ d_2 }{\lambda^2} \sqrt{\sum_{j = 1}^{d_2} C_2^2 d_1^2 ||W_{\cdot, j}- W'_{\cdot, j} ||^2_2} \quad \text{(by lemma~\ref{lemma_distance})}\\
\leq& \frac{ C_2 d_1 d_2 }{\lambda^2} ||W-W'||_F\\
\leq& \frac{ C_2 d_1 d_2 k^{1/2}}{\lambda^2} ||W-W'||_2
\end{align*}
where $h_j = \EE_{\x_i} \{  || \x_i||_2^2  |\mathrm{sign}(\langle \x_i, W_{\cdot, j} \rangle) -\mathrm{sign}(\langle \x_i, W'_{\cdot, j} \rangle)|^2 \}||_2$.

Thus we have
\begin{align*}
    ||\EE\{ Z_{i}^\top Z_{i} \}||_2 \leq \frac{ C_2 d_1 d_2 k^{1/2}}{\lambda^2} ||W-W'||_2 + ||W-W'||_2^2
\end{align*}
If $||W-W'||_2 < 1$, then $||W-W'||_2 \geq ||W-W'||_2^2$. And if $||W-W'||_2 > 1$, then $||W-W'||_2 \leq ||W-W'||_2^2$. Thus the above can be rewritten as
\begin{align*}
 ||\EE\{ Z_{i}^\top Z_{i} \}||_2 \leq \frac{ C_2 d_1 d_2 k^{1/2}}{\lambda^2} \max \{||W-W'||_2, ||W-W'||_2^2 \}
\end{align*}

\noindent \textbf{Bound $||\sum_{i=1}^m \EE\{ Z_{i} Z_{i}^\top \}||_2$}

Similarly, since $||\sum_{i=1}^m \EE\{Z_{i}  Z_{i}^\top \}||_2 \leq \sum_{i=1}^m ||\EE\{ Z_{i} Z_{i}^\top \}||_2$, we just need to bound $||\EE\{ Z_{i} Z_{i}^\top  \}||_2$.

\begin{align*}
&||\EE\{ Z_{i} Z_{i}^\top \}||_2\\
=&||\EE\{ (B_{i}-\EE B_{i})  (B_{i}-\EE B_{i})^\top \}||_2\\
=&||\EE\{ B_{i} B_{i}^\top -\EE B_{i} \cdot B_{i}^\top - B_{i} \cdot \EE B_{i}^\top + \EE B_{i} \EE B_{i}^\top\}||_2\\
=&||\EE\{ B_{i} B_{i}^\top \} - \EE B_{i} \EE B_{i}^\top||_2\\
\leq&||\EE\{ B_{i} B_{i}^\top \}||_2 + ||\EE B_{i} \EE B_{i}^\top||_2
\end{align*}

Therefore, we bound $||\EE\{ B_{i} B_{i}^\top \}||_2$ and $||\EE B_{i} \EE B_{i}^\top||_2$ respectively.

$$
||\EE B_{i} \EE B_{i}^\top||_2 = ||W-W'||^2_2
$$

\begin{align*}
&||\EE\{ B_{i} B_{i}^\top \}||_2 \\
=&\frac{d_2}{\lambda^2}||\EE\{  \x_i \bar{\e}_i^\top (\x_i \bar{\e}_i^\top)^\top |\mathrm{sign}(\langle \x_i \bar{\e}_i^\top, W \rangle) -\mathrm{sign}(\langle \x_i \bar{\e}_i^\top, W' \rangle)|^2 \}||_2 \\
=& \frac{d_2}{\lambda^2}||\EE\{  || \bar{\e}_i ||_2^2 \x_i \x_i^\top |\mathrm{sign}(\langle \x_i \bar{\e}_i^\top, W \rangle) -\mathrm{sign}(\langle \x_i \bar{\e}_i^\top, W' \rangle)|^2 \}||_2 \\
=& \frac{d_2}{\lambda^2}||\EE_{\x_i} \EE_{\bar{\e}_i}\{  || \bar{\e}_i ||_2^2 \x_i \x_i^\top |\mathrm{sign}(\langle \x_i \bar{\e}_i^\top, W \rangle) -\mathrm{sign}(\langle \x_i \bar{\e}_i^\top, W' \rangle)|^2 \}||_2 \\
=& \frac{d_2}{\lambda^2}||\frac{d_2}{d_2}  \sum_j \EE_{\x_i} \{ \x_i \x_i^\top |\mathrm{sign}(\langle \x_i , W_{\cdot,j} \rangle) -\mathrm{sign}(\langle \x_i , W'_{\cdot,j} \rangle)|^2 \}||_2 \\
\leq& \frac{d_2}{\lambda^2} \sum_j C_1 ||W_{\cdot, j} - W'_{\cdot, j}||_2 \quad \text{(by lemma~\ref{lemma_distance})}\\
\leq& \frac{d_2}{\lambda^2} \sqrt{d_2} C_1 ||W - W'||_F \\
\leq& \frac{ C_1 d_2^{3/2} k^{1/2}}{\lambda^2} ||W - W'||_2 
\end{align*}

Thus we have
\begin{align*}
    ||\EE\{  Z_{i} Z_{i}^\top \}||_2 \leq \frac{ C_1 d_2^{3/2} k^{1/2}}{\lambda^2} ||W-W'||_2 + ||W-W'||_2^2
\end{align*}
If $||W-W'||_2 < 1$, then $||W-W'||_2 \geq ||W-W'||_2^2$. And if $||W-W'||_2 > 1$, then $||W-W'||_2 \leq ||W-W'||_2^2$. Thus the above can be rewritten as
\begin{align*}
||\EE\{  Z_{i} Z_{i}^\top \}||_2 \leq \frac{ C_1 d_2^{3/2} k^{1/2}}{\lambda^2}  \max \{||W-W'||_2, ||W-W'||_2^2 \}
\end{align*}

And we can apply the Matrix Bernstein inequality to finish the final proof.
\end{proof}